%% file: acl2023.tex
\pdfoutput=1
\documentclass[11pt]{article}
\usepackage{authblk}
\usepackage[final]{ACL2023}
\usepackage{times}
\usepackage{latexsym}
\usepackage[T1]{fontenc}
\usepackage[utf8]{inputenc}
\usepackage{microtype}
\usepackage{inconsolata}
\usepackage{times}
\usepackage{xcolor}
\usepackage{graphicx}
\usepackage{latexsym}
\usepackage{array}
\usepackage{booktabs}
\usepackage{pgfplots}
\usepackage{algorithm}
\usepackage{algpseudocode}
\usepackage{graphicx}
\usepackage[T1]{fontenc}
\usepackage[utf8]{inputenc}
\usepackage{microtype}
\usepackage{xcolor}
\usepackage{soul}
\usepackage{fullpage}
\usepackage{enumitem}
\usepackage{pgfplots}
\usepackage{algorithm}
\usepackage{algpseudocode}
\usepackage{graphicx}
\usepackage{placeins}
\usepackage{tabularx}
\usepackage{makecell}
\usepackage{booktabs}       % professional-quality tables
\usepackage{array}          % tables with fixed lengths - enables using "m" to center content of non-multirow cells
\usepackage{colortbl}
\newcolumntype{P}[1]{>{\centering\arraybackslash}p{#1}}
\newcolumntype{M}[1]{>{\centering\arraybackslash}m{#1}}

\usepackage{subcaption}
\usepackage{amssymb}
\usepackage{amsmath}
\usepackage{mathtools}
\title{oBERTa: Improving Sparse Transfer Learning via improved initialization, distillation, and pruning regimes }%\thanks{~~~Corresponding author: dcampos3@illinois.edu} }
\author[1]{Daniel Campos } 
\author[2]{Alexandre Marques}
\author[2]{Mark Kurtz}
\author[1]{ChengXiang Zhai}
\affil[1]{Department of Computer Science, the University of Illinois Urbana-Champaign}
\affil[2]{Neural Magic Inc.}

\begin{document}
\maketitle
\begin{abstract}
In this paper, we introduce the range of oBERTa language models, an easy-to-use set of language models which allows Natural Language Processing (NLP) practitioners to obtain between 3.8 and 24.3 times faster models without expertise in model compression. Specifically, oBERTa extends existing work on pruning, knowledge distillation, and quantization and leverages frozen embeddings, improves distillation, and model initialization to deliver higher accuracy on a broad range of transfer tasks. In generating oBERTa, we explore how the highly optimized RoBERTa differs from the BERT for pruning during  pre-training and fine-tuning. We find it less amenable to compression during  fine-tuning. We explore the use of oBERTa on seven representative NLP tasks and find that the improved compression techniques allow a pruned oBERTa model to match the performance of BERT\textsubscript{base} and exceed the performance of Prune OFA Large on the SQUAD V1.1 Question Answering dataset, despite being 8x and 2x respectively faster in inference. We release our code, training regimes, and associated model for broad usage to encourage usage and experimentation. \footnote{https://github.com/neuralmagic/sparseml/}\textsuperscript{,}\footnote{https://sparsezoo.neuralmagic.com/}
\end{abstract}
\section{Introduction}
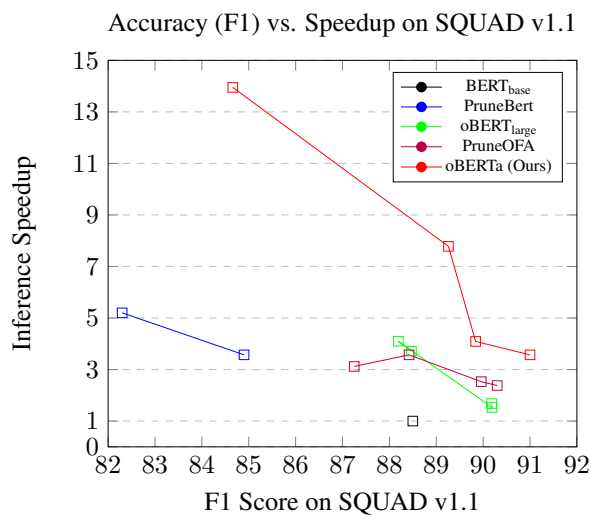
\begin{figure}[!htb]
\begin{tikzpicture}
\scalebox{0.9}{
\begin{axis}[
    title={Accuracy (F1) vs. Speedup on SQUAD v1.1},
    xlabel={ F1 Score on SQUAD v1.1},
    ylabel={Inference Speedup},
    xmin=82, xmax=92,
    ymin=0 , ymax=15,
    xtick={82,83,84,85,86,87,88, 89, 90, 91,92},
    ytick={0,1,3,5,7,9,11,13,15},
    legend pos=north east,
    ymajorgrids=true,
    grid style=dashed,
    legend style={nodes={scale=0.65, transform shape}}, 
    legend image post style={mark=*}
]
\addplot[
    color=black,
    mark=square,
    ]
    coordinates {
    (88.5,1)
    };
\addplot[
    color=blue,
    mark=square,
    ]
    coordinates {
    (84.9,3.57)(82.3,5.2)
    };
\addplot[
    color=green,
    mark=square,
    ]
    coordinates {
    (88.47,3.71)(88.19,4.1)(90.19,1.53)(90.18, 1.68)
    };
\addplot[
    color=purple,
    mark=square,
    ]
    coordinates {
    (87.25,3.12)(88.42,3.57)(89.96, 2.53)(90.30, 2.38)
    };
\addplot[
    color=red,
    mark=square,
    ]
    coordinates {
    (91.0,3.57)(89.84,4.09)(89.26,7.78)(84.66,13.95)
    };
%\legend{BERT-\textsubscript{base} \cite{Devlin2019BERTPO},PruneBert \cite{Sanh2020MovementPA},  oBERT\textsubscript{large} \cite{Frantar2021EfficientMA},  PruneOFA\cite{Zafrir2021PruneOF}, oBERTa (Ours)}
\legend{BERT\textsubscript{base} ,PruneBert,  oBERT\textsubscript{large} ,  PruneOFA, oBERTa (Ours)}
 \end{axis}}
\end{tikzpicture}
    \centering
    \caption{Performance of Sparse Language Models on the SQUAD V1.1 \cite{Rajpurkar2016SQuAD10} compared to an uncompressed BERT\textsubscript{base} \cite{Devlin2019BERTPO} with relation to realized inference improvements with regards to mean latency with a batch size of 1.}
    \label{fig:squadv1}
\end{figure}
The massive improvement in contextual word representations driven by the usage of the Transformer architecture \cite{Vaswani2017AttentionIA} has led to the wide-scale deployment of language models. These models are customized for various use cases and tasks like question answering, sentiment analysis, information retrieval, and document classification and deployed into general domains and specialized domains such as financial, medical, and legal. While these models are effective, they commonly contain hundreds of millions of parameters, which can lead to slow inference times without using specialized hardware accelerations like graphics processing units (GPU) or Tensor Processing Units (TPU). Without hardware acceleration, the inference on CPUs can be slow and impractical for real-world deployments.\\
Approaches such as knowledge distillation (KD) \cite{Hinton2015DistillingTK}, quantization \cite{Zafrir2019Q8BERTQ8}, and pruning \cite{Kurtic2022TheOB} have been leveraged to improve model efficiency and, when paired with specialized inference engines\footnote{https://github.com/neuralmagic/deepsparse}, it is possible to accelerate inference times on CPUs and GPUs significantly. While there has been substantial effort to create effective methods for compression \cite{Jiao2020TinyBERTDB, Sun2020MobileBERTAC} and improved model performance \cite{Liu2019RoBERTaAR}, general users of language models have been slower to adopt these methods. Years after its release, the original  BERT\textsubscript{base} uncased \cite{Devlin2019BERTPO} is still the most popular language model \footnote{Based on monthly downloads on the huggingface model hub in march 2023}, followed by the slightly compressed DistilBERT \cite{Sanh2019DistilBERTAD} for latency-sensitive deployments.
To enable broad adoption, regular users must be able to leverage more efficient language models without additional compression steps or tuning.\\
We present a case study on how to compress a language model for efficient CPU inference leveraging KD, structured pruning, unstructured sparsity, and quantization such that the compressed models can be applied to a broad range of natural language processing (NLP) tasks without expertise in compression of language models. \\
As part of this study, we release a set of efficient language models optimized to deliver the greatest improvement in inference while minimizing losses in accuracy. We then show how these models can be used for \textit{sparse transfer learning} \cite{Iofinova2021HowWD, Zafrir2021PruneOF} such that most compression happens during the pre-training stage. The pre-trained sparse models can be transferred to various NLP tasks, preserving sparsity without extensive optimization. Using these sparse transfer models and the DeepSparse inference engine, we show these sparse models can be fine-tuned to produce task-specific sparse models with minimal accuracy loss and result in greatly improved inference speeds with minimal accuracy loss.\\
As shown in Figure \ref{fig:squadv1}, oBERTa provides state-of-the-art performance for sparse language models on the SQUAD v1.1 Question Answering dataset. oBERTa variants exceed the performance of BERT\textsubscript{base} despite being eight times faster, exceed the performance of Prune OFA\textsubscript{large} and oBERT\textsubscript{large} while being two to five times faster.
In this paper, we focus on the following research questions: 
\begin{itemize}
    \item RQ1: Is RoBERTa more sensitive to unstructured pruning than BERT?
    \item RQ2: What is the impact of using a larger teacher for KD during the pruning of language models? 
    \item RQ3: Can frozen embeddings improve the accuracy of pruned language models?
\end{itemize}As part of our experimentation, we release the associated models and the training regimes to aid reproducibility and encourage efficient inference models. \\
In summary, our contributions are as follows:
\begin{itemize}
    \item We provide a thorough case study on how to compress a less studied language model\footnote{While the RoBERTa model was downloaded over 10m times in May 2023 on the huggingface hub it has not a model of focus for compression research.}, RoBERTa \cite{Liu2019RoBERTaAR}, and evaluate performance on a set of seven NLP tasks finding that it is possible to effectively compress a language model without using its original pre-training dataset.
    \item We demonstrate the impact of varying the size of teachers in KD, freezing embeddings, and variations in learning rates when applied to sparse language models.
    \item We demonstrate that our compressed models can be leveraged to deliver accuracy of over 91\% on the popular SQUAD v1.1 \cite{Rajpurkar2016SQuAD10} Question Answering Task with nearly three times faster inference than the previous state-of-the-art uses of unstructured sparsity.
\end{itemize} 
\section{Background and Related work}
While many methods to improve model efficiency exist, the same goal generally underpins them: given an original model $\theta$ with an accuracy of $acc(\theta)$ and an inference cost of $c(\theta)$ minimize the inference cost. While the methods used for compression can be highly optimized and specialized, they can commonly be used together to deliver massive improvements in inference speeds with minimal losses in accuracy. \\
\textbf{Transformer Based Language Models} such as BERT \cite{Devlin2019BERTPO} and T5 \cite{Raffel2020ExploringTL} provide contextual language representations built on the Transformer architecture \cite{Vaswani2017AttentionIA} which can be specialized and adapted for specific tasks and domains \cite{Lee2020BioBERTAP}. Using these models, it becomes relatively easy to excel at a broad range of natural language processing tasks such as Question Answering, Text Classification, and sentiment analysis. \\
\textbf{Unstructured Pruning} is a compression approach that removes individual weights or groups of weights in a model by applying a mask or setting the weight values to 0. This compression approach has been broadly studied in computer vision \cite{Han2015ADN}, and many methods can remove 70\% or more of model weights with little to no loss in accuracy. Models pruned can be 20x smaller in terms of pure model size and, when paired with a sparsity-aware inference engine such as DeepSparse \cite{deepsparse}, provide 3-5x speedups in inference throughput. \\
Focused on language models, recent work has shown that it is possible to prune models during fine-tuning \cite{Sanh2020MovementPA} \cite{Kurti2022TheOB} or during pre-training \cite{Zafrir2021PruneOF} and transfer to novel domains \cite{Campos2022SparseBERTSM} and datasets. \\
\textbf{Structured Pruning} is a compression approach that removes fundamental structural components in a language model such as individual attention heads \cite{Voita2019AnalyzingMS} or entire model layers such as transformer encoders \cite{sanh2019distilbert}. Structural pruning has become one of the most popular methods for inference optimization as it is easy to estimate the speedups and implement.\\
\textbf{Freezing Embeddings}, as introduced by Devlin et al. \cite{Devlin2019BERTPO}, involves training the embedding layer of a language model and then toggling the ability to continue to optimize, or not, the values of in the embeddings as training continues. \\
\textbf{Knowledge Distillation} \cite{Hinton2015DistillingTK} is a training method where a model is not explicitly a compression method but a training method where a model, called the \textit{student} learns to emulate a \textit{teacher} model which is commonly larger or better performing. The loss extracted from the original training data in KD is augmented or replaced by KL divergence between the student and teacher model. \\
KD leverages the hardness parameter $h$ to control the mixture of regular and distillation loss (with a higher distillation favoring the KL divergence loss) and a temperature parameter $t$ to control the softness of the distribution. \\
As applied to language models, the approach has been used to improve the performance of structurally pruned language models resulting in models like DistilBERT \cite{sanh2019distilbert} and TinyBERT \cite{Jiao2020TinyBERTDB}. \\
\textbf{Quantization} reduces the precision for the model weights and activations to lower the computational requirements of model execution. While researchers have explored reducing representation to binary representations \cite{Pouransari2020LeastSB}, current hardware limits inference speedups to 8 or 4-bit representations. Quantization can be applied after the model is trained in a one-shot fashion, but this can lead to large losses in accuracy because of rounding errors. To avoid this pitfall, quantization is applied as quantization-aware training (QAT), where the forward pass of the model is simulated with lower precision. In contrast, the backward pass happens in full precision. By using QAT models, learn to be robust to rounding errors and can result in quantization having little to no loss in accuracy. In language models, research has produced quantized language models such as Q8BERT \cite{Zafrir2019Q8BERTQ8} and is commonly used in conjunction with structured and unstructured pruning \cite{Zafrir2021PruneOF} as a way of introducing compounding compression.\\
Additional approaches such as early exiting \cite{Xin2020DeeBERTDE} or token pruning \cite{Kim2021LearnedTP} have also improved inference efficiency. Still, the inference improvements can be very dataset dependent and, as a result, out of our experimentation frame. For a broader survey on compression approaches, we recommend Treviso et al. recent work \cite{Treviso2022EfficientMF}\\
\section{Improving Sparse Transfer Learning}
\input{method}
\section{Experimental Results}
\input{experiment}
\section{Discussion}
\input{discussion}
\section{Limitations}
While our approach is effective at compressing models, it is not the most efficient. In order to discover the most optimal compression approaches and evaluate their performance performed hundreds of experiments. As a result, scaling our approach to every novel language understanding language model is not tractable.
Another limitation of our work is we did not track the complete compute utilization of our entire experimentation process but we can provide some estimates. Experiments in pruning during fine-tuning leveraged a single V100 16 GB GPU and took approximately 14 hours per experiment. The pre-training of structurally pruned models with knowledge distillation required 4 A100 40GB GPUs for approximately 72 hours. Pruning during pre-training with Knowledge distillation required approximately 100 hours on the same setup. Task-specific fine-tuning happened on a single V100 16GB GPU and depending on the size of the task was anywhere from a few minutes to 20 hours. Based on all of our experiments we estimate 400 V100 hours of pruning during fine-tuning, roughly 16,000 A100 hours\footnote{4000 hours and 4 A100 GPUS per hour} for pretraining, and assuming an average of 10 V100 hours per sparse transfer run, a total of 4000 V100 hours for sparse-transfer and sparse-transfer with quantization. 
\section{Conclusion and Future Work}
\bibliography{anthology,custom}
\bibliographystyle{acl_natbib}
\appendix
\section{Appendix}
\input{appendix}
\end{document}

%% file: method.tex
While quantization and pruning have been well studied as applied to language models, work has studied the compression BERT\textsubscript{base} or BERT\textsubscript{large}. Despite existing research, we find that a clear case study that explores how best to create a family of compressed models is lacking, and this work seeks to remedy that. As part of our research, we compare the impact of varying pruning methods, pruning stage, teachers for KD, and freezing portions of the model as applied to the RoBERTa language model.\\
While performing task-specific compression allows NLP practitioners to broadly adopt improvements in inference efficiency, having access to pre-optimized models is key. We produce a family of 8 general purpose language models, collectively called oBERTa, which progressively get smaller and faster with minimal losses in accuracy. \\
The oBERTa models leverage a combination of structured and unstructured pruning to provide a set of compressed models which can meet a wide set of latency needs. This compression approach has not been extensively documented nor discussed. Our approach to producing the oBERTA models builds on prior explorations of the combination of compression methods \cite{Kurti2022TheOB} and addresses compression approaches in a staged manner as shown in Figure \ref{fig:framework}.\\
First, we create three structural variants starting with a RoBERTa\textsubscript{base} model. The base uses 12 transformer layers, the medium uses 6, and the small uses 3. Following prior work, we select interleaved layers for the 6-layer model and the first, middle, and last layers for the 3-layer model. Then, each of these 3 models is further pre-trained using masked language modeling on the Wikipedia-Bookcorpus text dataset, leveraging KD from a  RoBERTa\textsubscript{large} teacher. After that, each model is pruned using gradual magnitude pruning (GMP) to a desired sparsity level (90\% and 95\%) during additional pre-training based on masked language modeling, similar to Zafir et al. \cite{Zafrir2021PruneOF}. Further background on the RoBERTA model and why we did not prune using the WebText corpus can be found in the appendix. \\
After pre-training, the sparsity profile is fixed, and models are fine-tuned and quantized on their target task with a small set of variable hyperparameters. Experimentation on the impact of larger teachers, frozen embeddings, and variations in pruning algorithms are discussed in subsequent portions of this work. 
\subsection{Downstream Compression}
We explore the impact of introducing unstructured sparsity during task-specific fine-tuning. We repeat each experiment with three different seeds and report the average F1 and Exact Match (EM) metrics in tables ~\ref{tab:OBS-downstream-squad} and \ref{tab:gmp-downstream-squad}. Following a basic hyperparameter sweep, our baseline RoBERTa\textsubscript{base} model achieves a performance of 83.95 EM and 91.13 F1 in the broadly used question-answering benchmark SQUAD V1.1 \cite{Rajpurkar2016SQuAD10}. \\
We also perform unstructured pruning varying the sparsity 50-95\% and the pruning method: GMP and Optimal BERT Surgeon (OBS) \cite{Kurti2022TheOB}. We prune each model for eight epochs, followed by an additional two epochs to allow the network to stabilize and re-converge. Knowledge distillation is used during training with the dense baseline model as a teacher, hardness set to $1.0$ and temperature set to $5.0$.  Further hyperparameters are in the appendix \ref{sec:downstream}.\\
Table \ref{tab:Sparse-Benchmark} shows the impact of sparsity on BERT\textsubscript{base}, as reported by previous work. Comparing these results with tables \ref{tab:OBS-downstream-squad} and \ref{tab:gmp-downstream-squad}, we conclude that RoBERTa is more sensitive to pruning than BERT, although RoBERTa\textsubscript{base} pruned with OBS remains substantially more accurate than BERT\textsubscript{base} for the same level of sparsity.\\
Table~\ref{tab:OBS-downstream-squad} shows that pruning RoBERTA\textsubscript{base} to 90\% with OBS results in a relative drop in F1 of 1.59\%, which is three times the relative drop reported for BERT\textsubscript{base} with the same pruning algorithm.
Moreover, table~\ref{tab:gmp-downstream-squad} shows that RoBERTA\textsubscript{base} becomes very sensitive to pruning with GMP for sparsities above 85\%, with the relative drop in F1 increasing almost threefold between 85\% and 90\% sparsity.
We conjecture that RoBERTa is more sensitive to pruning than BERT because the latter is relatively under-trained \cite{Liu2019RoBERTaAR}, making the more optimized RoBERTa more sensitive to the loss in expressivity caused by pruning.
\begin{table}[!ht]
    \centering
    \tiny
    \begin{tabular}{|l|l|l|l|}
    \hline
        Model & Sparsity& F1& Impact  \\ \hline
         BERT\textsubscript{base} \cite{Devlin2019BERTPO} & 0 & 88.50 & N/A \\ \hline
         BERT\textsubscript{large} \cite{Devlin2019BERTPO} & 0 &  90.9 & N/A \\ \hline
         RoBERTa\textsubscript{base} \cite{Liu2019RoBERTaAR} & 0 & 91.13 & N/A \\ \hline
         RoBERTA\textsubscript{large} \cite{Liu2019RoBERTaAR} & 0 & 94.60 & N/A \\ \hline
         PruneBert\textsubscript{base} \cite{Sanh2020MovementPA} & 90 & 84.90 & -4.07 \%\\ \hline
         PruneOFA\textsubscript{large} \cite{Zafrir2021PruneOF}& 90 & 87.25 & -1.41 \% \\ \hline
         oBERT\textsubscript{large} \cite{Kurti2022TheOB} & 90 & 87.98 &  -0.58\%  \\ \hline
         $GMP_\star$\textsubscript{large} \cite{Kurtic2022GMPWG} & 90 & 86.7 & -2.03\% \\ \hline
    \end{tabular}
    \caption{Performance of existing dense and sparse language models on the SQUAD v1.1 Question Answering Dataset}
    \label{tab:Sparse-Benchmark}
\end{table}
\begin{table}[!ht]
    \centering
    \small
    \begin{tabular}{|l|l|l|l|l|}
    \hline
        Sparsity (\%) & EM & Impact & F1 & Impact \\ \hline
        50 & 84.80 & 1.01\% & 91.49 & 0.40\% \\ \hline
        60 & 84.64 & 0.82\% & 91.33 & 0.22\% \\ \hline
        70 & 84.42 & 0.56\% & 91.13 & 0.00\% \\ \hline
        80 & 84.64 & 0.82\% & 91.33 & 0.22\% \\ \hline
        85 & 82.89 & -1.26\% & 90.12 & -1.11\% \\ \hline
        90 & 82.48 & -1.75\% & 89.68 & -1.59\% \\ \hline
        95 & 79.01 & -5.89\% & 87.05 & -4.47\% \\ \hline
    \end{tabular}
    \caption{Impact of Sparsity introduced by OBS on the F1 and EM scores of pruned RoBERTa models on the SQUAD V1.1 Dataset}
    \label{tab:OBS-downstream-squad}
\end{table}

\begin{table}[!ht]
    \centering
    \small
    \begin{tabular}{|l|l|l|l|l|}
    \hline
        Sparsity (\%) & EM & Impact & F1 & Impact \\ \hline
        50 & 84.90 & 1.13\% & 91.46 & 0.36\% \\ \hline
        60 & 84.27 & 0.38\% & 90.91 & -0.24\% \\ \hline
        70 & 83.37 & -0.69\% & 90.30 & -0.91\% \\ \hline
        80 & 81.64 & -2.76\% & 88.86 & -2.49\% \\ \hline
        85 & 81.64 & -2.76\% & 88.86 & -2.49\% \\ \hline
        90 & 76.51 & -8.86\% & 84.90 & -6.83\% \\ \hline
        95 & 69.39 & -17.34\% & 79.35 & -12.93\% \\ \hline
    \end{tabular}
    \caption{Impact of Sparsity introduced by GMP on the F1 and EM scores of pruned RoBERTa models on the SQUAD V1.1 Dataset}
    \label{tab:gmp-downstream-squad}
\end{table}
\subsection{Upstream Compression}
Based on our fine-tuning experiments, achieving a high degree of sparsity on the RoBERTA model leads to improvements in performance, but there are greater than expected losses in accuracy. Additionally, such compression is task-specific and non-amortizable, so we explore how best to generate general pruned RoBERTa models. While we eventually apply the winning set of training combinations to all of our variants of oBERTa, we first seek to answer the following questions: Does GMP or OBS perform better during pretraining pruning? Does Freezing the Embeddings during pretraining pruning further improve performance? Does the use of larger teachers further improve performance? \\
We prune various models while varying individual variables during pretraining to evaluate these questions. We experiment by pruning an oBERTa\textsubscript{base} (12 layers) model to 90\% and 95\% sparsity on all non-embedding layers. All pretraining pruning happens using the Wikipedia-BookCorpus dataset, where we train for five epochs using a learning rate of 5e-5 and a batch size of 256 using 4 A100 GPUS. To evaluate the impact of these models, we evaluate performance on the previously used SQUAD v1.1 question-answering dataset, where we train with a fixed training regime of 10 epochs with a learning rate of 1.5e-4 based on the work of Kurtic et al. We train without KD for each finetuning run with an unpruned RoBERTa\textsubscript{base} or an unpruned RoBERTa\textsubscript{large}. Details for the hyperparameters used to train all teacher models can be found in the appendix \ref{sec:TeacherStats}.  \\
\begin{table}[!ht]
    \centering
    \tiny
    \scalebox{0.65}{
    \begin{tabular}{|l|*4l|*4l|}
    \toprule
         &  \multicolumn{4}{l}{GMP} & \multicolumn{4}{l}{OBS} \\ 
        \toprule
        Model & F1 & Impact & EM & Impact & F1 & Impact & EM & Impact \\ \hline
        RoBERTA\textsubscript{base} & 92.18 & 0.00\% & 85.59 & 0.00\% & 92.18 & 0.00\% & 85.59 & 0.00\% \\ \hline
        oBERTa 90\% No KD & 88.34 & -4.17\% & 80.19 & -6.31\% & 87.72 & -4.83\% & 79.35 & -7.29\% \\ \hline
        oBERTa 90\% RoBERTA\textsubscript{base} KD & 88.75 & -3.72\% & 81.35 & -4.95\% & 88.60 & -3.88\% & 81.37 & -4.93\% \\ \hline
        oBERTa 90\% RoBERTA\textsubscript{large} KD & 89.65 & -2.75\% & 83.12 & -2.88\% & 89.63 & -2.76\% & 82.94 & -3.09\% \\ \hline
        oBERTa 95\% No KD & 86.58 & -6.07\% & 78.81 & -7.92\% & 84.90 & -7.90\% & 76.82 & -10.25\% \\ \hline
        oBERTa 95\% RoBERTA\textsubscript{base} KD & 86.99 & -5.63\% & 79.41 & -7.22\% & 86.14 & -6.55\% & 78.63 & -8.13\% \\ \hline
        oBERTa 95\% RoBERTA\textsubscript{large} KD & 87.60 & -4.97\% & 80.44 & -6.01\% & 86.14 & -6.55\% & 79.84 & -6.72\% \\ \hline
        \bottomrule
    \end{tabular}}
    \caption{Impact on F1 of SQUAD V1.1 of using OBS vs. GMP as the pruning method during pretraining. Impact measures the relative loss in performance vs. the unpruned RoBERTa\textsubscript{base} baseline.}
    \label{tab:OBS-vs-GMP-hard1.0}
\end{table}
Comparing the use of OBS vs. GMP as shown in table \ref{tab:OBS-vs-GMP-hard1.0}, we can see that GMP consistently outperforms OBS. This is the opposite of what is seen when pruning downstream or, in prior work, pruning BERT. Without access to the original training corpus OBS is likely unable to leverage the loss aware saliency importance as well as it can when it has the original dataset.  \\
\begin{table}[!ht]
    \centering
    \small
    \scalebox{0.57}{
    \begin{tabular}{|l|*4l|*4l|}
    \toprule
         &  \multicolumn{4}{l}{Hardness 0.5} & \multicolumn{4}{l}{Hardness 1.0} \\ 
        \toprule
        Model & F1 & Impact & EM & Impact & F1 & Impact & EM & Impact \\ \hline
         RoBERTA\textsubscript{base} & 92.18 & 0.00\% & 85.59 & 0.00\% & 92.18 & 0.00\% & 85.59 & 0.00\% \\ \hline
        oBERTa 90\% No KD & 88.21 & -4.31\% & 80.19 & -6.31\% & 88.34 & -4.17\% & 80.19 & -6.31\% \\ \hline
        oBERTa 90\% Base KD & 89.19 & -3.25\% & 81.74 & -4.50\% & 88.75 & -3.72\% & 81.35 & -4.95\% \\ \hline
        oBERTa 90\% Large KD & 90.14 & -2.21\% & 83.51 & -2.43\% & 89.65 & -2.75\% & 83.12 & -2.88\% \\ \hline
        oBERTa-95 No KD & 85.82 & -6.90\% & 77.77 & -9.14\% & 86.58 & -6.07\% & 78.81 & -7.92\% \\ \hline
        oBERTa-95 Base KD & 86.98 & -5.64\% & 79.23 & -7.43\% & 86.99 & -5.63\% & 79.41 & -7.22\% \\ \hline
        oBERTa-95 Large KD & 87.66 & -4.91\% & 80.40 & -6.07\% & 87.60 & -4.97\% & 80.44 & -6.01\% \\ \hline
        \bottomrule
    \end{tabular}}
    \caption{Impact on F1 of SQUAD V1.1 by hardness in KD during pretraining pruning. Impact measures the relative loss in performance vs. the unpruned RoBERTa\textsubscript{base} baseline.}
    \label{tab:hardness-oberta}
\end{table}
  Evaluating the impact of variations in the hardness  of KD as shown in table \ref{tab:hardness-oberta}, there is a bit more of a muted set of conclusions. The 95\% sparse models perform better with a hardness of 1.0, while the 90\% models do better with a hardness of 0.5. Given that our goal is to preserve most of the RoBERTa model without actually using its large dataset, we set our hardness to 1.0 as it keeps the model from explicitly learning the new dataset. \\
\begin{table}[!ht]
    \centering
    \small
    \scalebox{0.5}{
    \begin{tabular}{|l|*4l|*4l|}
    \toprule
          &  \multicolumn{4}{l}{Frozen Embeddings} & \multicolumn{4}{l}{Trained Embeddings} \\ 
        \midrule
        Model & F1 & Impact & EM & Impact & F1 & Impact & EM & Impact \\ \hline
        RoBERTa\textsubscript{base} & 92.18 & 0.00\% & 85.59 & 0.00\% & 92.18 & 0.00\% & 85.59 & 0.00\% \\ \hline
        oBERTa\textsubscript{base} 90\% no KD & 87.71 & -4.85\% & 79.62 & -6.98\% & 88.21 & -4.31\% & 80.19 & -6.31\% \\ \hline
        oBERTa\textsubscript{base} 90\% RoBERTa\textsubscript{base} KD & 89.7 & -2.69\% & 81.74 & -4.50\% & 89.19 & -3.24\% & 83.07 & -2.94\% \\ \hline
        oBERTa\textsubscript{base} 90\% RoBERTa\textsubscript{large} KD & 89.59 & -2.81\% & 82.98 & -3.05\% & 90.14 & -2.21\% & 83.51 & -2.43\% \\ \hline
        \bottomrule
    \end{tabular}}
    \caption{Impact on F1 of SQUAD V1.1 concerning the use of frozen embeddings or not during pretraining pruning. Impact measures the relative loss in performance vs. the unpruned RoBERTa\textsubscript{base} baseline.}
    \label{tab:freeze-embd-oberta}
\end{table}
When we evaluate the impact of freezing embeddings during pre-training, as shown in table \ref{tab:freeze-embd-oberta}, we find strong evidence that using frozen embeddings consistently leads to worse performance and, as a result, does not freeze embeddings during our model pruning. Looking at the impact of varying the size of the teacher for pretraining KD as shown in table \ref{tab:upsteam-teach}, we unsurprisingly find clear evidence that using a larger teacher during pretraining pruning leads to improvements in performance. \\
\begin{table}[!ht]
    \centering
    \small
    \scalebox{0.55}{
    \begin{tabular}{|l|*4l|*4l|}
    \toprule
          &  \multicolumn{4}{l}{Base Upstream Teacher} & \multicolumn{4}{l}{Large Upstream Teacher} \\ 
        \midrule
        Model & F1 & Impact & EM & Impact & F1 & Impact & EM & Impact \\ \hline
        RoBERTA\textsubscript{base} & 92.18 & 0.00\% & 85.59 & 0.00\% & 92.18 & 0.00\% & 85.59 & 0.00\% \\ \hline
        oBERTa 90\% no KD  & 88.34 & -4.17\% & 80.59 & -5.84\% & 88.1 & -4.43\% & 80.06 & -6.46\% \\ \hline
        oBERTa 90\% Base KD & 88.75 & -3.72\% & 81.35 & -4.95\% & 89.22 & -3.21\% & 82.02 & -4.17\% \\ \hline
        oBERTa 90\% Large KD & 89.65 & -2.74\% & 83.12 & -2.89\% & 89.98 & -2.39\% & 83.14 & -2.86\% \\ \hline
        \bottomrule
    \end{tabular}}
    
    \caption{Impact on F1 of SQUAD V1.1 with respect variation is the size of the teacher in KD during pretraining pruning. Impact measures the relative loss in performance vs. the unpruned RoBERTa\textsubscript{base} baseline.}
    \label{tab:upsteam-teach}
\end{table}
Using these experiments, we generate the recipe, which we then use to create the many variants of oBERTa. We evaluate their performance in Table \ref{tab:pretrain-numbers} where it is important to note that these results are accuracy, loss, and perplexity relative to the RoBERTa-large teacher, not the true dataset. The compression recipe, as shown in Figure \ref{fig:framework} is as follows: \\
\begin{enumerate}
    \item Starting with a pre-trained language model, removing some portion of transformer layers in an interleaved fashion. 
    \item Using Knowledge Distillation from a large uncompressed model,  pre-train the pruned model with a hardness of 1.0 and without freezing embeddings.
    \item Using Knowledge Distillation from a large uncompressed model, prune during further pretraining using GMP where sparsity levels are enforced at the parameter level. The resulting model is the sparse-transfer-student. 
    \item Train an uncompressed large language model on the desired NLP task's dataset. This is the sparse-transfer teacher. 
    \item Using the sparse-transfer teacher fine-tune  the sparse-transfer-student with knowledge distillation to convergence. Experiment with the use of frozen embeddings and various sizes of sparse-transfer teachers.
    \item Using the fine-tuned sparse-transfer student and teacher, train with quantization-aware training. If embeddings were frozen during initial fine-tuning they should be unfrozen here.  
\end{enumerate}

%% file: experiment.tex
Based on the aforementioned experiments, we generate 8 variants of oBERTa, each with a different size and sparsity profile; details can be found in table \ref{tab:oberta-model-info}. Within this table, we report the impact on the model size as measured by the raw and compressed size of the ONNX \footnote{https://onnx.ai/} model file. Embeddings are unpruned and each layer is pruned to the target sparsity profile independent of the rest of the model. As a result, the overall sparsity profile may vary as modules in the network may not be able to reach exactly 90\% or 95\% sparsity. \\
Using these \textit {inference-optimized} models, we evaluate their \textit{sparse transfer} performance by finetuning these models on their target task using a fixed training regime and minor hyperparameter exploration. For each task, we train them for 10 epochs or 20 (10 of which are Quantization Aware Training), with the longer schedule being reserved for models which are being quantized.  \\
We evaluate performance on a benchmark of diverse NLP tasks ranging from question answering, sentiment analysis, document classification, token classification, and text classification. For question answering, we leverage the SQuAD v1.1 \cite{Rajpurkar2016SQuAD10} and SQuAD V2.0 \cite{Rajpurkar2018KnowWY} datasets. We leverage the SST-2 \cite{socher-etal-2013-recursive} dataset for sentiment analysis. For text classification, we use the Quora Duplicate Query Detection (QQP) \cite{sambitsekhar_2017} and the MNLI \cite{N18-1101} datasets. We leverage the IMDB \cite{maas-EtAl:2011:ACL-HLT2011} dataset for document classification and CONLL2003 \cite{tjong-kim-sang-de-meulder-2003-introduction} for token classification.\\ 
Looking at performance on question answering as shown in table \ref{tab:sparse-transfer-squad} and \ref{tab:sparse-transfer-squad2}.
\begin{table}[!ht]
    \centering
    \tiny
    \scalebox{0.8}{
    \begin{tabular}{|l|*3l|*3l|}
    \toprule
         &  \multicolumn{3}{l}{Sparse Transfer} & \multicolumn{3}{l}{Sparse Transfer With Quantization} \\ 
        \midrule
        model & F1 & Recovery & EM & F1 & Recovery & EM \\ \hline
        oBERTa\textsubscript{base} & 92.15 & 100.00\% & 85.78 & 93.18 & 101.11\% & 87.29 \\ \hline
        oBERTa\textsubscript{base} 90\% & 90.95 & 98.69\% & 84.42 & 89.46 & 97.08\% & 82.61 \\ \hline
        oBERTa\textsubscript{base} 95\% & 89.84 & 97.49\% & 83.08 & 89.23 & 96.83\% &  81.12\\ \hline
        oBERTa\textsubscript{MEDIUM} & 90.37 & 98.06\% & 83.84 & 83.77 & 90.91\% & 90.37 \\ \hline
        oBERTa\textsubscript{MEDIUM} 90\% & 89.26 & 96.86\% & 82.18 & 88.65 & 96.20\% & 81.88 \\ \hline
        oBERTa\textsubscript{SMALL} & 84.87 & 92.09\% & 76.55 & 84.82 & 92.05\% & 76.77 \\ \hline
        oBERTa\textsubscript{SMALL} 90\% & 84.66 & 91.87\% & 76.18 & 82.18 & 92.18\% & 74.21\\ 
        \bottomrule
    \end{tabular}}
   
    \caption{Sparse Transfer performance of the oBERTA family on the SQUAD V1.1 dataset. The sparse transfer was performed over 10 epochs and sparse transfer with quantization over 20. Recovery is based on the relative performance of the unpruned oBERTa\textsubscript{base}. }
    \label{tab:sparse-transfer-squad}
\end{table}
\begin{table}[!ht]
    \centering
    \begin{tiny}
    \scalebox{0.8}{
    \begin{tabular}{|l|*3l|*3l|}
    \toprule
         &  \multicolumn{3}{l}{Sparse Transfer} & \multicolumn{3}{l}{Sparse Transfer With Quantization} \\ 
        \midrule
         model & F1 & Recovery & EM & F1 & Recovery & EM \\ \hline
        oBERTa\textsubscript{base} & 82.77 & 100.00\% & 79.56 & 85.298 & 103.06\% & 82.347 \\ \hline
        oBERTa\textsubscript{base} 90\% & 81.33 & 98.26\% & 78.27 & 81.43 & 98.38\% & 78.92 \\ \hline
        oBERTa\textsubscript{base} 95\% & 77.98 & 94.22\% & 74.67 & 78.09 & 94.35\% & 74.82 \\ \hline
        oBERTa\textsubscript{MEDIUM} & 77.51 & 93.65\% & 74.25 & 78.137 & 94.41\% & 75.179 \\ \hline
        oBERTa\textsubscript{MEDIUM} 90\% & 76.64 & 92.60\% & 73.34 & 76.24 & 92.11\% & 73.51 \\ \hline
        oBERTa\textsubscript{SMALL} & 71.54 & 86.44\% & 67.93 & 71.591 & 86.50\% & 68.087 \\ \hline
        oBERTa\textsubscript{SMALL} 90\% & 70.79 & 85.53\% & 67.31 & 69.35 & 87.79\% & 65.21 \\ \hline
        \bottomrule
    \end{tabular}}
    \end{tiny}
    \caption{Sparse Transfer performance of the oBERTA family on the SQUAD V2.0 dataset. The sparse transfer was performed over 10 epochs, and sparse transfer with quantization over 20. Recovery is based on the relative performance of the unpruned oBERTa\textsubscript{base}. }
    \label{tab:sparse-transfer-squad2}
\end{table}
Moving to text classification on QQP and MNLI as shown in tables \ref{tab:sparse-transfer-qqp} and \ref{tab:sparse-transfer-MNLI}
\begin{table}[!ht]
    \centering
    \begin{tiny}
    \scalebox{0.70}{
    \begin{tabular}{|l|*3l|*3l|}
    \toprule
         &  \multicolumn{3}{l}{Sparse Transfer} & \multicolumn{3}{l}{Sparse Transfer With Quantization} \\ 
        \midrule
         model & Accuracy & Recovery &Accuracy(MM) & Accuracy & Recovery & Accuracy(MM) \\ \hline
        oBERTa\textsubscript{base} & 87.88\% & 100.00\% & 87.57\% & 88.06\% & 100.20\% & 88.01\% \\ \hline
        oBERTa\textsubscript{base} 90\% & 85.17\% & 96.91\% & 84.73\% & 85.09\% & 96.83\% & 84.76\% \\ \hline
        oBERTa\textsubscript{base} 95\% & 84.32\% & 95.95\% & 84.08\% & 83.73\% & 95.28\% & 83.83\% \\ \hline
        oBERTa\textsubscript{MEDIUM} & 85.29\% & 97.05\% & 85.17\% & 83.62\% & 95.15\% & 83.74\% \\ \hline
        oBERTa\textsubscript{MEDIUM} 90\% & 81.61\% & 92.87\% & 81.32\% & 82.37\% & 93.73\% & 81.79\% \\ \hline
        oBERTa\textsubscript{SMALL} & 80.80\% & 91.95\% & 81.55\% & 81.10\% & 92.29\% & 81.51\% \\ \hline
        oBERTa\textsubscript{SMALL} 90\% & 79.23\% & 90.15\% & 79.24\% & 79.14\% & 90.06\% & 79.42\% \\ \hline
        \bottomrule
    \end{tabular}}
    \end{tiny}
    \caption{Sparse Transfer performance of the oBERTA family on the MNLI dataset. Sparse transfer was performed over 10 epochs and sparse transfer with quantization over 20. Recovery is based on the relative performance of the unpruned oBERTa\textsubscript{base}. }
    \label{tab:sparse-transfer-MNLI}
\end{table}
\begin{table}[!ht]
    \centering
    \begin{tiny}
    \scalebox{0.6}{
    \begin{tabular}{|l|*4l|*4l|}
    \toprule
         &  \multicolumn{4}{l}{Sparse Transfer} & \multicolumn{4}{l}{Sparse Transfer With Quantization} \\ 
        \midrule
        model & Accuracy & Recovery & F1 & Combined & Accuracy & Recovery & F1 & Combined \\ \hline
        oBERTa\textsubscript{base} & 91.52\% & 100.00\% & 90.09\% & 88.66\% & 89.86\% & 98.18\% & 88.12\% & 86.73\% \\ \hline
        oBERTa\textsubscript{base} 90\% & 91.01\% & 99.44\% & 89.47\% & 87.92\% & 91.21\% & 99.66\% & 89.68\% & 88.16\% \\ \hline
        oBERTa\textsubscript{base} 95\% & 90.85\% & 99.26\% & 89.21\% & 87.58\% & 90.72\% & 99.12\% & 89.08\% & 0.87\% \\ \hline
        oBERTa\textsubscript{MEDIUM} & 91.35\% & 99.81\% & 89.90\% & 88.44\% & 91.33\% & 99.79\% & 89.80\% & 88.28\% \\ \hline
        oBERTa\textsubscript{MEDIUM} 90\% & 90.48\% & 98.86\% & 88.85\% & 87.21\% & 90.60\% & 99.00\% & 89.01\% & 87.42\% \\ \hline
        oBERTa\textsubscript{SMALL} & 90.72\% & 99.13\% & 89.21\% & 87.71\% & 89.74 & 98.06\% & 87.99 & 86.25 \\ \hline
        oBERTa\textsubscript{SMALL} 90\% & 89.74\% & 98.06\% & 87.99\% & 86.25\% & 89.73 & 98.04\% & 87.98 & 86.08\\
        \bottomrule
    \end{tabular}}
    \end{tiny}
    \caption{Sparse Transfer performance of the oBERTA family on the QQP dataset. The sparse transfer was performed over ten epochs, and sparse transfer with quantization over 20. Recovery is based on the relative performance of the unpruned oBERTa\textsubscript{base}.}
    \label{tab:sparse-transfer-qqp}
\end{table}
Shifting focus to document classification as shown in table \ref{tab:sparse-transfer-imdb} and sentiment analysis in \ref{tab:sparse-transfer-SST2}
\begin{table}[!ht]
    \centering
    \begin{tiny}
    \scalebox{0.95}{
    \begin{tabular}{|l|*2l|*2l|}
    \toprule
         &  \multicolumn{2}{l}{Sparse Transfer} & \multicolumn{2}{l}{Sparse Transfer With Quantization} \\ 
        \midrule
        model & Accuracy & Recovery & Accuracy & Recovery \\ \hline
        oBERTa\textsubscript{base} & 95.24\% & 100.00\% & 95.44\% & 100.21\% \\ \hline
        oBERTa\textsubscript{base} 90\% & 93.64\% & 98.32\% & 93.28 & 97.94\% \\ \hline
        oBERTa\textsubscript{base} 95\% & 93.48\% & 98.15\% & 92.80& 97.23\% \\ \hline
        oBERTa\textsubscript{MEDIUM} & 93.36\% & 98.03\% & 94.08 & 98.78\% \\ \hline
        oBERTa\textsubscript{MEDIUM} 90\% & 92.24\% & 96.85\% & 92.08& 96.69\% \\ \hline
        oBERTa\textsubscript{SMALL} & 93.04\% & 97.69\% & 92.52 & 97.15\% \\ \hline
        oBERTa\textsubscript{SMALL} 90\% & 91.60\% & 96.18\% & 91.28 & 95.84\% \\
        \bottomrule
    \end{tabular}}
    \end{tiny}
    \caption{Sparse Transfer performance of the oBERTA family on the IMDB dataset. The sparse transfer was performed over ten epochs, and sparse transfer with quantization over 20. Recovery is based on the relative performance of the unpruned oBERTa\textsubscript{base}. }
    \label{tab:sparse-transfer-imdb}
\end{table}
\begin{table}[!ht]
    \centering
    \begin{tiny}
    \scalebox{0.95}{
    \begin{tabular}{|l|*2l|*2l|}
    \toprule
         &  \multicolumn{2}{l}{Sparse Transfer} & \multicolumn{2}{l}{Sparse Transfer With Quantization} \\ 
        \midrule
        model & Accuracy & Recovery & Accuracy & Recovery \\ \hline
        oBERTa\textsubscript{base} & 94.60 & 100.00\% & 92.66 & 97.95\% \\ \hline
        oBERTa\textsubscript{base} 90\% & 92.78 & 98.08\% & 92.546 & 97.83\% \\ \hline
        oBERTa\textsubscript{base} 95\% & 91.51 & 96.74\% & 91.399 & 96.62\% \\ \hline
        oBERTa\textsubscript{MEDIUM} & 92.89 & 98.19\% & 91.06 & 96.26\% \\ \hline
        oBERTa\textsubscript{MEDIUM} 90\% & 88.76 & 93.83\% & 89.91 & 95.04\% \\ \hline
        oBERTa\textsubscript{SMALL} & 90.48 & 95.64\% & 91.28 & 96.49\% \\ \hline
        oBERTa\textsubscript{SMALL} 90\% & 89.34 & 94.44\% & 88.65 & 93.71\% \\ 
        \bottomrule
    \end{tabular}}
    \end{tiny}
    \caption{Sparse Transfer performance of the oBERTA family on the SST-2 dataset. The sparse transfer was performed over ten epochs, and sparse transfer with quantization over 20. Recovery is based on the relative performance of the unpruned oBERTa\textsubscript{base}. }
    \label{tab:sparse-transfer-SST2}
\end{table}
Finally, looking at performance on token classification as shown in table \ref{tab:sparse-transfer-conll2003} 
\begin{table}[!ht]
    \centering
    \begin{tiny}
    \scalebox{0.75}{
    \begin{tabular}{|l|*3l|*3l|}
    \toprule
         &  \multicolumn{3}{l}{Sparse Transfer} & \multicolumn{3}{l}{Sparse Transfer With Quantization} \\ 
        \midrule
        model & Accuracy & Recovery & F1 & Accuracy & Recovery & F1 \\ \hline
        oBERTa\textsubscript{base} & 99.26\% & 100.00\% & 95.51\% & 99.30\% & 100.05\% & 95.98\% \\ \hline
        oBERTa\textsubscript{base} 90\% & 99.11\% & 99.85\% & 94.98\% & 99.05\% & 99.79\% & 94.51\% \\ \hline
        oBERTa\textsubscript{base} 95\% & 98.89\% & 99.63\% & 93.32\% & 98.75\% & 99.48\% & 92.61\% \\ \hline
        oBERTa\textsubscript{MEDIUM} & 99.04\% & 99.77\% & 94.39\% & 99.18\% & 99.92\% & 95.15\% \\ \hline
        oBERTa\textsubscript{MEDIUM} 90\% & 98.79\% & 99.53\% & 93.31\% & 98.73\% & 99.46\% & 92.70\% \\ \hline
        oBERTa\textsubscript{SMALL} & 99.01\% & 99.75\% & 94.00\% & 98.98\% & 99.72\% & 94.13\% \\ \hline
        oBERTa\textsubscript{SMALL} 90\% & 98.47\% & 99.20\% & 91.13\% & 98.25\% & 98.98\% & 89.79\% \\ 
        \bottomrule
    \end{tabular}}
    \end{tiny}
    \caption{Sparse Transfer performance of the oBERTA family on the CONLL-2003 dataset. The sparse transfer was performed over ten epochs, and sparse transfer with quantization over 20. Recovery is based on the relative performance of the unpruned oBERTa\textsubscript{base}. }
    \label{tab:sparse-transfer-conll2003}
\end{table}

\subsection{Inference Benchmark}
To evaluate the performance of our inference-optimized models, we benchmark performance using the popular DeepSparse library version 1.3.2 \footnote{pip install deepsparse==1.3.2} and an Intel Xeon Gold 6238R Processor. Performance is measured using models that have been \textit{sparse-transferred} to the SQuAD v1.1 dataset and exported to a standard ONNX model format. Benchmarks are run on 4 and 24 cores and a sequence length of 384 with batch sizes of 1, 16, and 64. For each model, the benchmark is run for 60 seconds with a warm-up period of 10 seconds, and we report the throughput (items per second) and the mean, median, and standard deviation per item latency. 
\begin{table}[!ht]
    \centering
    \begin{tiny}
    \scalebox{0.85}{
    \begin{tabular}{|l|*3l|*3l|}
    \toprule
         &  \multicolumn{3}{l}{24 Cores} & \multicolumn{3}{l}{4 Cores} \\ 
        \midrule
        Model & BS 1 & BS 16 & BS 64 &  BS 1 & BS 16 & BS 64 \\
        \midrule
        BERT\textsubscript{base} & 1.00 & 1.00 & 1.00 & 1.00 & 1.00 & 1.00 \\ \hline
        oBERTa\textsubscript{base} & 1.00 & 1.00 & 1.00 & 1.00 & 1.00 & 1.00 \\ \hline
        oBERTa\textsubscript{base} Quantized & 3.10 & 4.29 & 4.46 & 4.09 & 4.31 & 4.32 \\ \hline
        oBERTa\textsubscript{base} 90\% & 3.29 & 3.80 & 3.80 & 3.60 & 3.34 & 3.40 \\ \hline
        oBERTa\textsubscript{base} 90\% Quantized & 4.12 & 7.05 & 7.37 & 7.67 & 7.59 & 7.40 \\ \hline
        oBERTa\textsubscript{base} 95\% & 8.72 & 4.56 & 4.65 & 4.12 & 3.85 & 4.37 \\ \hline
        oBERTa\textsubscript{base} 95\% Quantized & 4.73 & 8.22 & 8.56 & 9.41 & 9.06 & 8.68 \\ \hline
        oBERTa\textsubscript{MEDIUM} & 1.96 & 1.99 & 1.99 & 1.96 & 1.99 & 2.02 \\ \hline
        oBERTa\textsubscript{MEDIUM} Quantized & 6.20 & 8.04 & 8.44 & 8.43 & 8.33 & 8.45 \\ \hline
        oBERTa\textsubscript{MEDIUM} 90\%   & 6.35 & 7.41 & 6.84 & 7.83 & 6.56 & 6.72 \\ \hline
        oBERTa\textsubscript{MEDIUM} 90\% Quantized  & 8.94 & 12.86 & 13.65 & 14.99 & 14.81 & 14.95 \\ \hline
        oBERTa\textsubscript{SMALL} & 3.89 & 3.96 & 3.99 & 3.95 & 3.97 & 4.03 \\ \hline
        oBERTa\textsubscript{SMALL} Quantized & 12.47 & 14.12 & 14.08 & 15.50 & 15.48 & 15.70 \\ \hline
        oBERTa\textsubscript{SMALL} 90\% & 12.22 & 14.40 & 14.67 & 14.05 & 14.19 & 14.13 \\ \hline
        oBERTa\textsubscript{SMALL} 90\% Quantized & 16.21 & 21.35 & 23.96 & 29.77 & 27.14 & 27.58 \\ \hline
        \bottomrule
    \end{tabular}}
    \end{tiny}
    \caption{Latency reduction of the oBERTa family concerning the unpruned oBERTa\textsubscript{base} as measured on 24 and 4 cores. Speedup is measured relative to the latency reduction in MS/batch, and BS refers to batch size.}
    \label{tab:inference-bench}
\end{table}
We present a set of summary statistics of relative speedup across batch sizes and inference server configurations as shown in table \ref{tab:inference-bench}. Full inference performance results can be found in the appendix. In analyzing performance, we can see that the introduction of quantization to a dense model delivers roughly a 4x speedup while quantization on sparse models is closer to 2x. With the introduction of sparsity, 90\% leads to slightly under 4x speedup, while 95\% leads to slightly over 4x. The impact of structural pruning is roughly equivalent to the size of the as a 6-layer model is two times faster than a 12-layer, and a 3-layer model is four times faster. Combing compression forms is only partially additive, as a small (3-layer) 90\% quantized model performance is ~24x vs the expected 32x (4x from structural pruning, 2x quantization, 4x unstructured pruning. \\
Looking at the variation in a speedup by batch size and the number of cores, we can see that allocating more cores leads to a smaller gap in inference speedup, especially with small batches. From this, we extract that compression is significant when performing streaming inference (batch size 1) on smaller CPUs.  \\
Next, we go ahead and benchmark the oBERTa model performance against existing sparse-transfer models such as oBERT and PruneOFA using the models that have been published \footnote{Since the PruneBERT model is not available in the zoo, we extrapolate numbers using the performance of our oBERTa\textsubscript{base} pruned 90\% as both models feature 12 transformer encoders and 90\% sparsity.} in Neural Magic's Sparse-Zoo \footnote{https://sparsezoo.neuralmagic.com/}. We run these models using four cores and a batch size of 1 and compare their speedup (or slowdown) relative to their performance on the SQUAD v1.1 question-answering benchmark. Results can be found in table \ref{tab:inference-competitive-short} and full results in \ref{tab:inference-competitive-full}. Looking at the improvements in accuracy and inference throughput, we find the oBERTa models are 1.3 to 4 times better than models with approximately the same accuracy.  \\
\begin{table}[!ht]
    \centering
    \scalebox{0.5}{
    \begin{tabular}{|l|l|*2l|*2l|}
    \toprule
        \multicolumn{2}{l}{} &  \multicolumn{2}{l}{Vs. BERT\textsubscript{base}} & \multicolumn{2}{l}{Vs. BERT\textsubscript{large}} \\ \hline
        Model & F1 & Recovery & Speedup & Recovery & Speedup \\ \hline
        oBERTa\textsubscript{base} 90\% & 91.00 & 102.77\% & 3.57 & 100.44\% & 20.21 \\ \hline
        oBERT\textsubscript{large} 95\% Quantized & 90.21 & 101.87\% & 3.41 & 99.57\% & 19.31 \\ \hline
        \midrule
        prunedOFA\textsubscript{large} 90\% Quantized & 89.96 & 101.59\% & 2.38 & 99.29\% & 13.47 \\ \hline
        oBERTa\textsubscript{base} 90\% Quantized & 89.46 & 101.03\% & 7.62 & 98.74\% & 43.07 \\ \hline
        \midrule
        oBERTa\textsubscript{MEDIUM} 90\%   & 89.26 & 98.99\% & 7.78 & 96.75\% & 43.99 \\ \hline
        obert\textsubscript{base} 90\% Quantized & 88.00 & 99.38\% & 6.96 & 97.13\% & 39.37 \\ \hline
        \midrule
        oBERTa\textsubscript{SMALL} 90\% & 84.66 & 90.97\% & 13.95 & 88.91\% & 78.91 \\ \hline
        pruneBERT 90\% & 84.90 & 95.88\% & 3.57  & 93.71\% & 73.82 \\ \hline
        \bottomrule
    \end{tabular}}
    \caption{Speedups of the oBERTa-family compared to existing published sparse models compared to the performance of BERT\textsubscript{base} and BERT-large. Speedup measures the reduction in latency of MS/batch. oBERTa\textsubscript{base} 90\% exceeds the accuracy of oBERT\textsubscript{large} 95\% quantized despite being faster, oBERTa\textsubscript{base} 90\% quantized performs at the level of pruneOFA\textsubscript{large} 90\% Quantized despite being 3x faster, oBERTa\textsubscript{MEDIUM} 90\% can outperform oBERT\textsubscript{base} 90\% Quantized despite being 30\% faster, and oBERTa\textsubscript{SMALL} 90\% performs on par with pruneBERT 90\% despite being nearly four times faster. }
    \label{tab:inference-competitive-short}
\end{table}
Looking at the competitive results, we find that the oBERTa-* models can deliver significant gains in performance (F1) relative to speedups. The oBERTa\textsubscript{base}Pruned 90\% Quantized model achieves an undertaking that nearly matches pruneOFA-large 90\% Quantized while delivering nearly 13x faster inference. Similarly, the oBERTA\textsubscript{SMALL} 90\%  model provides similar accuracy to PruneBERT despite being over four times faster.

%% file: discussion.tex
\textbf{Sparse Models require higher learning rates} as shown in the tables in \ref{sec:sparse-transfer-learning-rate} sparse language models can be used as general-purpose contextual language models but require the use of a much higher learning rate. When using structurally pruned models like the 6-layer oBERTa\textsubscript{MEDIUM} and the 3-layer oBERTa\textsubscript{SMALL}, the optimal learning rate does not vary much within the same task despite the model size. With the introduction of sparsity, the learning rate needs to scale, usually by a factor of five or ten. We find this counterintuitive as the sparse models have fewer parameters to \textit{tune}, so we would expect them to prefer a much lower learning rate. We attribute this to the loss of expressivity in the network driven by its sparsity. Since the network has fewer degrees of freedom to optimize the points which can be optimized move much more than those that cannot. \\
\textbf{Larger models compress better} as shown by the gap between the sparse and dense models and the gap between models and their quantized counterparts. While 12-layer models can receive 90 or 95 \% sparsity and quantization with little to no loss in accuracy, the three and 6-layer models see a much bigger dip. This aligns with Li et al. 2020 \cite{Li2020TrainLT} in which they demonstrate that larger models are more robust to pruning and quantization. Empirically, this makes sense as the smaller models have \textit{fewer degrees of freedom}, and other portions of the network cannot counteract the reduction in expressivity caused by pruning and quantization.   \\
\textbf{Bigger Teachers are not always better} as shown in the table in \ref{sec:sparse-transfer-KD} the introduction of larger teachers does not always lead to improvements in accuracy. The impact is highly task and model dependent as some datasets like MNLI or QQP see little impact in using larger teachers, yet datasets like SQUAD or SQUAD v2.0  see large impacts, which are even more pronounced when the student model is smaller. \\
\textbf{Frozen embeddings can help}, but not always. As shown by \ref{sec:sparse-transfer-freeze-embd} the impact of freezing the embeddings is highly task-specific and inconsistent across tasks or models. In question answering, freezing leads to 1-2 point movement for unpruned models and 5-7 points for pruned models. In other tasks like QQP and MNLI, the impact of frozen embeddings tends to be minor or none. 

%% file: appendix.tex
\label{sec:appendix}
\subsection{Model Generation Approach}
\begin{figure*}[!t]
    \centering
    \scalebox{0.7}{\includegraphics{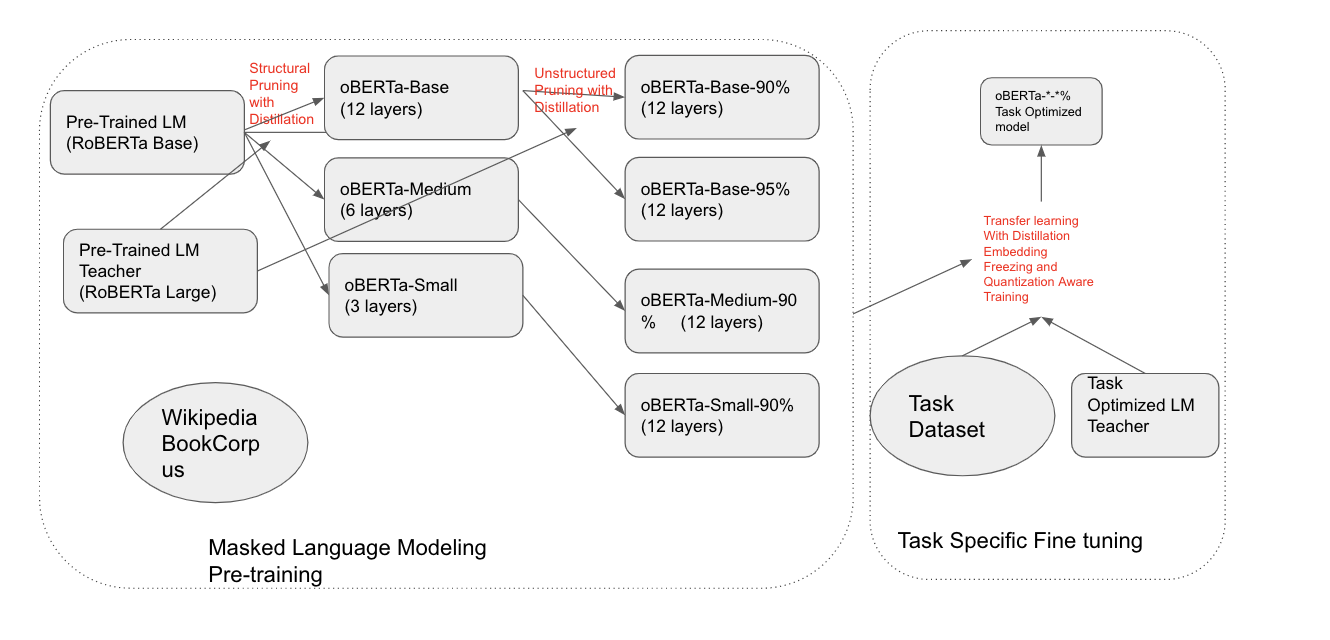}}
    \caption{The set of oBERTa language models follows a compounding compression approach. First models are structurally pruned and further pre-trained using KD and a RoBERTa\textsubscript{large}  teacher. Next, each model is pruned during additional pre-training to a target sparsity. After  pruning, the sparsity pattern is locked, and models are fine-tuned with KD on specialized NLP tasks. During fine-tuning, models may be quantized for additional improvements in inference efficiency.}
    \label{fig:framework}
\end{figure*}

oBERTa models are generated in a multi-stage approach with details found in figure \ref{fig:framework}
\subsection{Roberta and Training Methodology}
RoBERTa \cite{Liu2019RoBERTaAR} is a language model that can best be considered more robust and optimized for the popular BERT model. While the models share architectures, their training differs as RoBERTA uses a 160 GB corpus for 10 epochs compared to the 4GB one used by BERT. As a result, the training time of RoBERTA is about 100 times higher than its predecessor.\\
Given this high cost of training and the regular need for longer training when pruning a model \cite{Kurti2022TheOB}, we focus on compressing RoBERTa without following its expensive pre-training regime. Our research leverages the popular open-source compression library SparseML\footnote{https://github.com/neuralmagic/sparseml} to implement unstructured pruning, structured pruning, and quantization via quantization-aware training. In all our experiments, we prune each network component independently using either GMP or OBS (Kurtic et al.). One exception is the embeddings layer, which we do not prune.\\
\begin{table}[!ht]
    \centering
    \caption{Pretraining performance using knowledge distillation from a RoBERTa large model. }
    \begin{tabular}{|l|l|l|l|}
    \hline
        Model & ACC & Loss & Perplexity \\ \hline
        oBERTa\textsubscript{base}  & 0.580 & 3.775 & 43.593 \\ \hline
        oBERTa\textsubscript{base} 90\% & 0.506 & 4.448 & 85.420 \\ \hline
        oBERTa\textsubscript{base} 95\% & 0.439 & 4.734 & 113.702 \\ \hline
        oBERTa\textsubscript{medium} & 0.533 & 4.296 & 73.391 \\ \hline
        oBERTa\textsubscript{medium} 90\% & 0.631 & 1.896 & 6.662 \\ \hline
        oBERTa\textsubscript{small} & 0.465 & 4.561 & 95.670 \\ \hline
        oBERTa\textsubscript{small} 90\% & 0.404 & 4.669 & 106.614 \\ \hline
    \end{tabular}
    \label{tab:pretrain-numbers}
\end{table}
\subsection{Model Details}
Model details can be found in table \ref{tab:oberta-model-info}
\begin{table*}[!tb]
    \centering
    \scalebox{0.65}{
    \begin{tabular}{|l|l|l|l|l|l|l|l|l|}
    \hline
        Model & Parameters & Prunable & Sparse & Sparsity & size (MB) & Compression & GZIP size (MB) & Compression \\ \hline
        oBERTa\textsubscript{base} & 124,647,170 & 85,526,016 & 1,539 & 0.0\% & 474 & 1.00 & 435 & 1.00 \\ \hline
        oBERTa\textsubscript{base} Quantized & 124,647,170 & 85,526,016 & 1,539 & 0.0\% & 119 & 3.98 & 85 & 5.12 \\ \hline
        oBERTa\textsubscript{base} 90\% & 124,647,170 & 85,526,016 & 76,442,738 & 89.4\% & 474 & 1.00 & 183 & 2.38 \\ \hline
        oBERTa\textsubscript{base} 90\% Quantized & 124,647,170 & 85,526,016 & 76,442,738 & 89.4\% & 119 & 3.98 & 42 & 10.36 \\ \hline
        oBERTa\textsubscript{base} 95\% & 124,647,170 & 85,526,016 & 80,689,466 & 94.3\% & 474 & 1.00 & 163 & 2.67 \\ \hline
        oBERTa\textsubscript{base} 95\% Quantized & 124,647,170 & 85,526,016 & 80,689,466 & 94.3\% & 119 & 3.98 & 37 & 11.76 \\ \hline
        oBERTa\textsubscript{MEDIUM} & 82,119,938 & 43,058,688 & 1,538 & 0.0\% & 312 & 1.52 & 289 & 1.51 \\ \hline
        oBERTa\textsubscript{MEDIUM} Quantized & 82,119,938 & 43,058,688 & 1,538 & 0.0\% & 78 & 6.08 & 53 & 8.21 \\ \hline
        oBERTa\textsubscript{MEDIUM} 90\%   & 82,119,938 & 43,058,688 & 38,222,138 & 88.8\% & 312 & 1.52 & 161 & 2.70 \\ \hline
        oBERTa\textsubscript{MEDIUM} 90\% Quantized  & 82,119,938 & 43,058,688 & 38,222,138 & 88.8\% & 78 & 6.08 & 33 & 13.18 \\ \hline
        oBERTa\textsubscript{SMALL} & 60,856,322 & 21,825,024 & 1,538 & 0.0\% & 233 & 2.03 & 214 & 2.03 \\ \hline
        oBERTa\textsubscript{SMALL} Quantized & 60,856,322 & 21,825,024 & 1,538 & 0.0\% & 60 & 7.90 & 39 & 11.15 \\ \hline
        oBERTa\textsubscript{SMALL} 90\% & 60,856,322 & 21,825,024 & 19,111,068 & 87.6\% & 233 & 2.03 & 149 & 2.92 \\ \hline
        oBERTa\textsubscript{SMALL} 90\% Quantized & 60,856,322 & 21,825,024 & 19,111,838 & 87.6\% & 60 & 7.90 & 30 & 14.50 \\ \hline
    \end{tabular}}
    \caption{Description of the oBERTa model family and their sparsity and size. Prunable parameters are the sum of all non-embedding parameters in the model. Since sparsity profiles are assigned at a module level, overall sparsity profiles do not perfectly match the target 90\% or 95\% which are targeted.}
    \label{tab:oberta-model-info}
\end{table*}
\subsection{Dataset Details}
\label{sec:datasets}
Dataset statistics are detailed in Table \ref{tab:dataset_stat}.
\begin{table}[!htb]
      \centering
         {\small 
             \begin{tabular}{l|c|c}
                \toprule 
                Dataset & Train & Eval \\
                \midrule
                SQuAD v1.1 (examples) &  87599 & 10570\\
                SQuAD v2.0 (examples) &  130319	 & 11873\\
                \midrule
                MNLI (examples) & 392702 & 19628 \\
                \midrule
                QQP (examples) &  363,846 & 40,430 \\
                \midrule
                IMDB (examples) & 25000	 & 25000 \\
                \midrule
                CONLL2003 (examples) & 14041	 & 3250 \\
                \midrule
                SST2 (examples) & 67349	 & 872	 \\
                \midrule
                Wikipedia (words) & 6078422  & - \\
                \midrule
                TBC (words) & 74004228 & - \\ 
                \bottomrule
            \end{tabular}
         }
     \caption{Statistics for training and evaluation datasets}
    \label{tab:dataset_stat}
\end{table}
\subsection{Teacher models}
\label{sec:TeacherStats}
Performance of the RoBERTa\textsubscript{base}and RoBERTa\textsubscript{large} models on our sparse transfer datasets. We explore the optimal hyperparameters relative to performance in published results as shown in table \ref{tab:teacher-base} and \ref{tab:teacher-large}
\begin{table*}[!ht]
    \centering
    \scalebox{0.7}{
    \begin{tabular}{|l|l|l|l|l|l|l|l|l|l|l|}
    \hline
        Model & Training Epochs & Batch Size & Learning Rate & Weight Decay & Warmup & Target Metric & Target Score & Actual & Recall \\ \hline
        SQUAD V1.1 & 3 & 16 & 1.00E-05 & 0 & 0  & F1 & 90.40 & 92.15 & 101.94\% \\ \hline
        SQUAD V2.0  & 3 & 16 & 3.00E-05 & 0 & 0 & F1 & 82.91 & 83.53 & 100.74\% \\ \hline
        QQP & 5 & 16 & 2.00E-05 & 0 & 0 & ACC & 91.90 & 91.52 & 99.59\% \\ \hline
        MNLI & 3 & 16 & 1.00E-05 & 0 & 0 &  ACC & 87.60 & 87.88 & 100.31\% \\ \hline
        SST-2 & 3 & 16 & 2.00E-05 & 0 & 0 &  ACC & 94.80 & 94.61 & 99.80\% \\ \hline
        CONLL2003 & 3 & 16 & 3.00E-05 & 0 & 0 &  ACC & 99.10 & 99.29 & 100.19\% \\ \hline
        IMDB & 3 & 16 & 1.00E-05 & 0 & 0 &  ACC & 94.67 & 95.24 & 100.60\% \\ \hline
    \end{tabular}}
    \caption{Training parameters along with performance metrics and the recovery vs. the published performance of the same model for the RoBERTa base model}
    \label{tab:teacher-base}
\end{table*}
\begin{table*}[!ht]
    \centering
    \scalebox{0.7}{
    \begin{tabular}{|l|l|l|l|l|l|l|l|l|l|l|}
    \hline
        Model & Training Epochs & Batch Size & Learning Rate & Weight Decay & Warmup &Target Metric & Target Score & Actual & Recall \\ \hline
        SQUAD V1.1 & 3 & 16 & 1.00E-05 & 0 & 0  & F1 & 94.50 & 94.62 & 100.12\% \\ \hline
        SQUAD V2.0  & 3 & 16 & 1.00E-05 & 0 & 0  & F1 & 89.40 & 89.14 & 99.71\% \\ \hline
        QQP & 3 & 16 & 1.00E-05 & 0 & 0 &  ACC & 92.20 & 91.76 & 99.52\% \\ \hline
        MNLI & 3 & 16 & 1.00E-05 & 0 & 0 &  ACC & 90.20 & 90.61 & 100.45\% \\ \hline
        SST-2 & 3 & 16 & 1.00E-05 & 0 & 0 &  ACC & 96.40 & 96.22 & 99.81\% \\ \hline
        CONLL2003 & 3 & 16 & 3.00E-05 & 0 & 0 &  ACC & 99.10 & 99.39 & 100.29\% \\ \hline
        IMDB & 3 & 16 & 1.00E-05 & 0 & 0 &  ACC & 94.67 & 96.12 & 101.53\% \\ \hline
    \end{tabular}}
    \caption{Training parameters along with performance metrics and the recovery vs. the published performance of the same model for the RoBERTa large model}
    \label{tab:teacher-large}
\end{table*}
\subsection{Upstream Pruning}
Following the findings that more extensive teachers distill better \cite{Liu2019RoBERTaAR} and our experiments, we use both RoBERTa\textsubscript{base}and RoBERTa\textsubscript{large} as teachers eventually find the large model works better. Using this teacher, we use the parameters shown in table \ref{tab:hyperparams-UpstreamPruning} to prune the models for oBERTa. This same set of parameters is applied to the structurally pruned models, but there is no induced sparsity. 
\begin{table*}
      \centering
         {\small 
            \begin{tabular}{l|c}
            \toprule
            & 5 Epochs \\
            \midrule
            Datasets & BookCorpus \& English Wikipedia \\
            \midrule
            Batch size & 256 \\
            \midrule
            Initial learning rate & 5e-4 \\
            Learning rate schedule & linear decay with rewinds \\
            Learning rate rewinds & periodic every 0.5 epochs \\
            \midrule
            Max sequence length & 512 \\
            Weight decay & 0.01 \\
            \midrule
            \makecell{Knowledge Distillation\\(hardness, temperature)} & (1.0, 5.5) \\
            \midrule
            Student model & dense oBERTa-* model \\
            Teacher model & RoBERTa\textsubscript{large} \\
            \midrule
            Pruning frequency & 100x per epoch \\
            \bottomrule
            \midrule
            Initial Sparsity & 0.7 for 12 layer model, 0.5 for the 6-layer, and 0.3 for the 3-layer \\
            \bottomrule
            \end{tabular}
        }
            \caption{Upstream pruning hyper-parameters.}
            \label{tab:hyperparams-UpstreamPruning}
\end{table*}
\subsection{Sparse Transfer Hyper-parameters}
\label{sec:downstream}
Our work aims not to produce the highest possible performance of a sparse language model. Instead, we aim to make light language models that perform well on various tasks with minimal hyperparameter optimization. As a result, in all of our experiments, we leverage the parameters shown in \ref{tab:hyperparams-transfer} and \ref{tab:hyperparams-transfer-quant} and perform a grid search over them. 
\begin{table*}
      \centering
        {\small 
            \begin{tabular}{l|c}
            \toprule
            & 10 Epochs \\
            \midrule
            Initial learning rate & \makecell{2.1e-4,1.9e-4,1.7e-4,1.5e-4,1.3e-4,1.1e-4,9e-5,7e-5,5e-5,3e-5,2e-5,1e-5} \\
            Learning rate schedule & linear decay to 0 \\
            \midrule
                Batch size & 12 \\
            \midrule
            \midrule
                Weight Decay & \makecell{0.0, 0.01, 0.05, 0.1} \\
            \midrule
            \midrule
               Knowledge Distillation hardness & \makecell{1.0, 0.0} \\
            \midrule
            \midrule
               Frozen Embeddings & \makecell{1.0, 0.0} \\
            \midrule
            \midrule
               Knowledge Distillation temperature & 7.0 \\
            \midrule
            \midrule
               Knowledge Distillation Teacher & \makecell{RoBERTa\textsubscript{base}, RoBERTa\textsubscript{large}} \\
            \midrule
            \bottomrule
            \end{tabular}
        }
        \caption{Sparse-transfer learning hyper-parameters used to fine-tune upstream-pruned models at downstream tasks. Each Experiment tunes this set of parameters to find a task-specific optimal combination.}
    \label{tab:hyperparams-transfer}
\end{table*}
\begin{table*}
      \centering
        {\small 
            \begin{tabular}{l|c}
            \toprule
            & 20 Epochs \\
            \midrule
            Initial learning rate & \makecell{2.1e-4,1.9e-4,1.7e-4,1.5e-4,1.3e-4,1.1e-4,9e-5,7e-5,5e-5,3e-5,2e-5,1e-5} \\
            Learning rate schedule & linear decay to 0. Rewind to 5e-5 for QAT at epoch 10 \\
            \midrule
                Freeze Batch Norm Epoch & 18 \\
            \midrule
            \midrule
                Batch size & 12 \\
            \midrule
            \midrule
                Weight Decay & \makecell{0.0, 0.01, 0.05, 0.1} \\
            \midrule
            \midrule
               Knowledge Distillation hardness & \makecell{1.0, 0.0} \\
            \midrule
            \midrule
               Frozen Embeddings & \makecell{1.0, 0.0} \\
            \midrule
            \midrule
               Frozen Embeddings Schedule & Frozen until epoch 10, unfrozen for QAT \\
            \midrule
            \midrule
               Knowledge Distillation temperature & 7.0 \\
            \midrule
            \midrule
               Knowledge Distillation Teacher & \makecell{RoBERTa\textsubscript{base}, RoBERTa\textsubscript{large}} \\
            \midrule
            \bottomrule
            \end{tabular}
        }
        \caption{Sparse-transfer learning with Quantization hyper-parameters used to fine-tune upstream-pruned models at downstream tasks. Each Experiment tunes this set of parameters to find a task-specific optimal combination.}
    \label{tab:hyperparams-transfer-quant}
\end{table*}
\subsection{Learning Rate}
\label{sec:sparse-transfer-learning-rate}
In our exploration of sparse transfer learning, we perform a wide study on the impact of the optimal learning rate for each task and each model in the oBERTa family. The results as shown in table \ref{tab:learning-rate}
\begin{table*}
      \centering
        {\small 
        \begin{tabular}{|l|l|l|l|l|l|l|l|}
    \hline
         & \multicolumn{7}{l}{Optimal Learning Rate} \\ \hline
        model & SQUAD & SQUAD V2 & MNLI & QQP & IMDB & SST2 & CONLL2003 \\ \hline
        RoBERTa\textsubscript{base}& 1.00E-05 & 3.00E-05 & 1.00E-05 & 2.00E-05 & 1.00E-05 & 2.00E-05 & 3.00E-05 \\ \hline
        RoBERTa\textsubscript{large} & 1.00E-05 & 1.00E-05 & 1.00E-05 & 1.00E-05 & 1.00E-05 & 1.00E-05 & 3.00E-05 \\ \hline
        oBERTa\textsubscript{base}& 1.00E-05 & 1.00E-05 & 1.00E-05 & 2.00E-05 & 1.00E-05 & 2.00E-05 & 3.00E-05 \\ \hline
        oBERTa\textsubscript{base} 90\% & 1.50E-04 & 1.50E-04 & 7.00E-05 & 1.70E-04 & 1.30E-04 & 9.00E-05 & 1.50E-04 \\ \hline
        oBERTa\textsubscript{base} 95\% & 1.50E-04 & 1.30E-04 & 9.00E-05 & 2.10E-04 & 1.30E-04 & 9.00E-05 & 5.00E-05 \\ \hline
        oBERTa\textsubscript{MEDIUM} & 5.00E-05 & 5.00E-05 & 2.00E-05 & 3.00E-05 & 3.00E-05 & 2.00E-05 & 3.00E-05 \\ \hline
        oBERTa\textsubscript{MEDIUM} 90\%   & 1.50E-04 & 1.30E-04 & 1.50E-04 & 1.50E-04 & 5.00E-05 & 1.50E-04 & 1.50E-04 \\ \hline
        oBERTa\textsubscript{SMALL} & 1.50E-04 & 1.50E-04 & 3.00E-05 & 5.00E-05 & 3.00E-05 & 5.00E-05 & 3.00E-05 \\ \hline
        oBERTa\textsubscript{SMALL} 90\% & 1.50E-04 & 1.50E-04 & 2.10E-04 & 2.10E-04 & 1.50E-04 & 2.10E-04 & 1.90E-04 \\ \hline
    \end{tabular}    
        }
        \caption{Sparse-transfer learning with Quantization hyper-parameters used to fine-tune upstream-pruned models at downstream tasks. Each Experiment tunes this set of parameters to find a task-specific optimal combination.}
    \label{tab:learning-rate}
\end{table*}
\subsection{Knowledge Distillation}
\label{sec:sparse-transfer-KD}
In our exploration of sparse transfer learning, we perform a wide study on the impact of knowledge distillation. Across tasks, we look at the impact using no teacher, RoBERTa\textsubscript{base}and RoBERTa\textsubscript{large} as shown in tables \ref{tab:kd-mnli},\ref{tab:kd-qqp},\ref{tab:kd-sst},\ref{tab:kd-conll},\ref{tab:kd-squad},\ref{tab:kd-squad2}

\begin{table}[!htb]
    \centering
    \scalebox{0.7}{
    \begin{tabular}{|l|l|l|l|}
    \hline
        model & No KD & KD-Base & KD-Large \\ \hline
        oBERTa\textsubscript{base}(Target) & 91.52\% & N/A & N/A \\ \hline
        oBERTa\textsubscript{base} 90\% & 91.97 & 92.78 & 92.55 \\ \hline
        oBERTa\textsubscript{base} 95\% & 91.40 & 91.17 & 91.514 \\ \hline
        oBERTa\textsubscript{MEDIUM} & 90.94 & 91.86 & 91.78 \\ \hline
        oBERTa\textsubscript{MEDIUM} 90\% & 87.16 & 87.16 & 89.56 \\ \hline
        oBERTa\textsubscript{SMALL} & 89.56 & 88.65 & 90.83 \\ \hline
        oBERTa\textsubscript{SMALL} 90\% & 85.58 & 89.22 & 89.45 \\ \hline
    \end{tabular}}
    \caption{Impact of knowledge distillation on the accuracy (matched) MNLI Dataset across model sizes for the various sizes of oBERTa as compared to the regularly trained baseline}
    \label{tab:kd-mnli}
\end{table}

\begin{table}[!htb]
    \centering
    \scalebox{0.7}{
    \begin{tabular}{|l|l|l|l|}
    \hline
        model & No KD & KD-Base & KD-Large \\ \hline
        oBERTa\textsubscript{base}(Target) & 91.52 & N/A & N/A \\ \hline
        oBERTa\textsubscript{base} 90\% & 63.18 & 91.01 & 90.93 \\ \hline
        oBERTa\textsubscript{base} 95\% & 90.46 & 90.45 & 90.72 \\ \hline
        oBERTa\textsubscript{MEDIUM} & 90.75 & 90.96 & 90.96 \\ \hline
        oBERTa\textsubscript{MEDIUM} 90\% & 89.93 & 90.41 & 89.82 \\ \hline
        oBERTa\textsubscript{SMALL} & 86.63 & 87.34 & 87.65 \\ \hline
        oBERTa\textsubscript{SMALL} 90\% & 88.72 & 89.40 & 87.50 \\ \hline
    \end{tabular}}
    \caption{Impact of knowledge distillation on the accuracy QQP Dataset across model sizes for the various sizes of oBERTa as compared to the regularly trained baseline}
    \label{tab:kd-qqp}
\end{table}

\begin{table}[!htb]
    \centering
    \scalebox{0.7}{
    \begin{tabular}{|l|l|l|l|}
    \hline
        model & No KD & KD-Base & KD-Large \\ \hline
        oBERTa\textsubscript{base}(Target) & 91.52 & N/A & N/A \\ \hline
        oBERTa\textsubscript{base} 90\% & 91.97 & 92.78 & 92.55 \\ \hline
        oBERTa\textsubscript{base} 95\% & 91.4 & 91.17 & 91.514 \\ \hline
        oBERTa\textsubscript{MEDIUM} & 90.94 & 91.86 & 91.78 \\ \hline
        oBERTa\textsubscript{MEDIUM} 90\% & 87.16 & 87.16 & 89.56 \\ \hline
        oBERTa\textsubscript{SMALL} & 89.56 & 88.65 & 90.83 \\ \hline
        oBERTa\textsubscript{SMALL} 90\% & 85.58 & 89.22 & 89.45 \\ \hline
    \end{tabular}}
    \caption{Impact of knowledge distillation on the accuracy SST-2 Dataset across model sizes for the various sizes of oBERTa as compared to the regularly trained baseline}
    \label{tab:kd-sst}
\end{table}

\begin{table}[!htb]
    \centering
    \scalebox{0.7}{
    \begin{tabular}{|l|l|l|l|}
    \hline
        model & No KD & KD-Base & KD-Large \\ \hline
        oBERTa\textsubscript{base}(Target) & 91.52\% & N/A & N/A \\ \hline
        oBERTa\textsubscript{base} 90\% & 99.17 & 99.08 & 99.11 \\ \hline
        oBERTa\textsubscript{base} 95\% & 98.89 & 98.47 & 97.51 \\ \hline
        oBERTa\textsubscript{MEDIUM} & 99.21 & 99.16 & 99.19 \\ \hline
        oBERTa\textsubscript{MEDIUM} 90\% & 99.01 & 98.8 & 98.79 \\ \hline
        oBERTa\textsubscript{SMALL} & 99.05 & 98.95 & 98.94 \\ \hline
        oBERTa\textsubscript{SMALL} 90\% & 98.88 & 98.55 & 98.55 \\ \hline
    \end{tabular}}
    \caption{Impact of knowledge distillation on the accuracy on the CONLL2003 Dataset across model sizes for the various sizes of oBERTa as compared to the regularly trained baseline}
    \label{tab:kd-conll}
\end{table}

\begin{table}[!htb]
    \centering
    \scalebox{0.7}{
    \begin{tabular}{|l|l|l|l|}
    \hline
        model & No KD & KD-Base & KD-Large \\ \hline
        oBERTa\textsubscript{base}(Target) & 91.52\% & N/A & N/A \\ \hline
        oBERTa\textsubscript{base} 90\% & 89.01 & 90.86 & 90.92 \\ \hline
        oBERTa\textsubscript{base} 95\% & 87.06 & 89.84 & 89.21 \\ \hline
        oBERTa\textsubscript{MEDIUM} & 84.36 & 88.20 & 85.74 \\ \hline
        oBERTa\textsubscript{MEDIUM} 90\% & 84.71 & 89.26 & 88.61 \\ \hline
        oBERTa\textsubscript{SMALL} & 82.00 & 80.77 & 77.08 \\ \hline
        oBERTa\textsubscript{SMALL} 90\% & 73.31 & 84.66 & 83.13 \\ \hline
    \end{tabular}}
    \caption{Impact of knowledge distillation on the F1 SQUAD v1.1 Dataset across model sizes for the various sizes of oBERTa as compared to the regularly trained baseline}
    \label{tab:kd-squad}
\end{table}

\begin{table}[!htb]
    \centering
    \scalebox{0.7}{
    \begin{tabular}{|l|l|l|l|}
    \hline
         model & No KD & KD-Base & KD-Large \\ \hline
        oBERTa\textsubscript{base}(Target) & 91.52\% & N/A & N/A \\ \hline
        oBERTa\textsubscript{base} 90\% & 75.57852204 & 80.25256971 & 81.32561567 \\ \hline
        oBERTa\textsubscript{base} 95\% & 72.61 & 77.67 & 77.98 \\ \hline
        oBERTa\textsubscript{MEDIUM} & 69.42634 & 70.97328 & 71.55996 \\ \hline
        oBERTa\textsubscript{MEDIUM} 90\% & 68.25281 & 76.02975 & 76.64135 \\ \hline
        oBERTa\textsubscript{SMALL} & 66.8281 & 62.9573 & 63.1224 \\ \hline
        oBERTa\textsubscript{SMALL} 90\% & 55.3959 & 70.0796 & 70.7913 \\ \hline
    \end{tabular}}
    \caption{Impact of knowledge distillation on the F1 SQUAD v2.0 Dataset across model sizes for the various sizes of oBERTa as compared to the regularly trained baseline}
    \label{tab:kd-squad2}
\end{table}
\subsection{Freezing Embeddings}
\label{sec:sparse-transfer-freeze-embd}
In our exploration of sparse transfer learning, we perform a wide study on the impact of freezing the embeddings during finetuning. Across tasks, we look at the impact of frozen and unfrozen embeddings as shown in tables \ref{tab:mnli-freeze},\ref{tab:qqp-freeze},\ref{tab:sst2-freeze},\ref{tab:CONLL-freeze},\ref{tab:squad-freeze}, and \ref{tab:squad2-freeze}. Besides question answering, we do not find a strong trend with the impact of frozen embeddings. In some tasks, sparse and dense models perform better with frozen embeddings while not for others. Focusing on question answering, by using frozen embeddings dense models see large losses in F1 score and the opposite can be seen for pruned models.   
\begin{table}[!htb]
    \centering
    \scalebox{0.99}{
    \begin{tabular}{|l|l|l|}
    \hline
        model & Frozen & Unfrozen \\ \hline
        oBERTa\textsubscript{base} (Target) & N/A & 87.88\% \\ \hline
        oBERTa\textsubscript{base} 90\% & 84.50 & 83.81 \\ \hline
        oBERTa\textsubscript{base} 95\% & 83.91 & 83.41 \\ \hline
        oBERTa\textsubscript{MEDIUM} & 84.37 & 83.32 \\ \hline
        oBERTa\textsubscript{MEDIUM} 90\% & 81.61 & 77.00 \\ \hline
        oBERTa\textsubscript{SMALL} & 80.24 & 80.36 \\ \hline
        oBERTa\textsubscript{SMALL} 90\% & 78.46 & 74.25 \\ \hline
    \end{tabular}}
    \caption{Impact of frozen vs trained embeddings on the accuracy (matched) MNLI Dataset across model sizes for the various sizes of oBERTa as compared to the uncompressed baseline}
    \label{tab:mnli-freeze}
\end{table}
\begin{table}[!htb]
    \centering
    \begin{tabular}{|l|l|l|}
    \hline
        model & Frozen & Unfrozen \\ \hline
        oBERTa\textsubscript{base} (Target) & N/A & 91.52\% \\ \hline
        oBERTa\textsubscript{base} 90\% & 90.93\% & 90.99\% \\ \hline
        oBERTa\textsubscript{base} 95\% & 90.72\% & 90.85\% \\ \hline
        oBERTa\textsubscript{MEDIUM} & 90.96\% & 91.35\% \\ \hline
        oBERTa\textsubscript{MEDIUM} 90\% & 89.82\% & 90.48\% \\ \hline
        oBERTa\textsubscript{SMALL} & 90.59\% & 90.72\% \\ \hline
        oBERTa\textsubscript{SMALL} 90\% & 89.40\% & 89.74\% \\ \hline
    \end{tabular}
    \caption{Impact of frozen vs trained embeddings on the accuracy on QQP across model sizes for the various sizes of oBERTa as compared to the uncompressed baseline}
    \label{tab:qqp-freeze}
\end{table}
\begin{table}[!ht]
    \centering
    \begin{tabular}{|l|l|l|}
    \hline
    model & Frozen & Unfrozen \\ \hline   
    oBERTa\textsubscript{base} (Target) & N/A & 91.52\% \\ \hline
        oBERTa\textsubscript{base} 90\% & 92.55 & 91.74\\ \hline
        oBERTa\textsubscript{base} 95\% & 91.514 & 91.4 \\ \hline
        oBERTa\textsubscript{MEDIUM} & 91.78 & 92.89 \\ \hline
        oBERTa\textsubscript{MEDIUM} 90\% & 89.56 & 88.76 \\ \hline
        oBERTa\textsubscript{SMALL} & 90.83 & 90.48 \\ \hline
        oBERTa\textsubscript{SMALL} 90\% & 89.45 & 89.34 \\ \hline    
    \end{tabular}
    \caption{Impact of frozen vs trained embeddings on the accuracy SST2 Dataset across model sizes for the various sizes of oBERTa as compared to the uncompressed baseline}
    \label{tab:sst2-freeze}
\end{table}
\begin{table}[!ht]
    \centering
    \begin{tabular}{|l|l|l|}
    \hline
     model & Frozen & Unfrozen \\ \hline
        oBERTa\textsubscript{base} (Target) & N/A & 91.52\% \\ \hline
        oBERTa\textsubscript{base} 90\% & 97.51 & 98.55 \\ \hline
        oBERTa\textsubscript{base} 95\% & 99.11 & 99.13 \\ \hline
        oBERTa\textsubscript{MEDIUM} & 99.19 & 99.18 \\ \hline
        oBERTa\textsubscript{MEDIUM} 90\% & 98.79 & 98.9 \\ \hline
        oBERTa\textsubscript{SMALL} & 98.94 & 98.94 \\ \hline
        oBERTa\textsubscript{SMALL} 90\% & 98.55 & 98.69 \\ \hline
    \end{tabular}
    \caption{Impact of frozen vs trained embeddings on the accuracy on CONLL2003 Dataset across model sizes for the various sizes of oBERTa as compared to the uncompressed baseline}
    \label{tab:CONLL-freeze}
\end{table}

\begin{table}[!ht]
    \centering
    \begin{tabular}{|l|l|l|}
    \hline
        model & Frozen & Unfrozen \\ \hline
        oBERTa\textsubscript{base} (Target) & N/A & 91.52\% \\ \hline
        oBERTa\textsubscript{base} 90\% & 90.92 & 83.99 \\ \hline
        oBERTa\textsubscript{base} 95\% & 89.21 & 87.08 \\ \hline
        oBERTa\textsubscript{MEDIUM} & 85.74 & 89.95 \\ \hline
        oBERTa\textsubscript{MEDIUM} 90\% & 88.61 & 86.63 \\ \hline
        oBERTa\textsubscript{SMALL} & 77.08 & 84.64 \\ \hline
        oBERTa\textsubscript{SMALL} 90\% & 83.13 & 77.43 \\ \hline
    \end{tabular}
    \caption{Impact of frozen vs trained embeddings on SQUAD v1.1 F1 across model sizes for the various sizes of oBERTa as compared to the uncompressed baseline}
    \label{tab:squad-freeze}
\end{table}

\begin{table}[!ht]
    \centering
    \begin{tabular}{|l|l|l|}
    \hline
        model & Frozen & Unfrozen \\ \hline
        oBERTa\textsubscript{base} (Target) & N/A & 91.52\% \\ \hline
        oBERTa\textsubscript{base} 90\% & 71.56 & 78.05 \\ \hline
        oBERTa\textsubscript{base} 95\% & 81.33 & 78.45 \\ \hline
        oBERTa\textsubscript{MEDIUM} & 77.98 & 76.86 \\ \hline
        oBERTa\textsubscript{MEDIUM} 90\% & 76.64 & 72.77 \\ \hline
        oBERTa\textsubscript{SMALL} & 71.32 & 63.12 \\ \hline
        oBERTa\textsubscript{SMALL} 90\% & 70.79 & 59.38 \\ \hline
    \end{tabular}
    \caption{Impact of frozen vs trained embeddings on the SQUAD v2.0 Dataset across model sizes for the various sizes of oBERTa as compared to the uncompressed baseline}
    \label{tab:squad2-freeze}
\end{table}
\subsection{Inference Benchmarks}
\label{sec:sparse-transfer-params}
We provide full results for our experiments in benchmarking the impact of compression on inference efficiency as shown in tables \ref{tab:inference-competitive-full},\ref{tab:inference-bs-64-core4},\ref{tab:inference-bs-16-core4},\ref{tab:inference-bs-1-core24},\ref{tab:inference-bs-64-core24},\ref{tab:inference-bs-16-core24},\ref{tab:inference-bs-1-core4-competitive},\ref{tab:inference-bs-1-core4-competitive}
\begin{table*}[!ht]
    \centering
    \scalebox{0.60}{
    \begin{tabular}{|l|l|l|l|l|l|}
    \hline
        model & Throughput (items/sec) & Speedup & Latency Mean (ms/batch) & Latency Median (ms/batch & Latency Std (ms/batch) \\ \hline
        oBERTa\textsubscript{base} & 16.69 & 1.00 & 59.90 & 59.82 & 1.02 \\ \hline
        oBERTa\textsubscript{base} Quantized & 51.68 & 3.10 & 19.34 & 19.28 & 0.58 \\ \hline
        oBERTa\textsubscript{base} 90\% & 54.87 & 3.29 & 18.21 & 18.15 & 0.31 \\ \hline
        oBERTa\textsubscript{base} 90\% Quantized & 68.70 & 4.12 & 14.55 & 14.50 & 0.20 \\ \hline
        oBERTa\textsubscript{base} 95\% & 145.57 & 8.72 & 6.86 & 6.86 & 0.11 \\ \hline
        oBERTa\textsubscript{base} 95\% Quantized & 78.90 & 4.73 & 12.66 & 12.68 & 0.31 \\ \hline
        oBERTa\textsubscript{MEDIUM} & 32.78 & 1.96 & 30.49 & 30.44 & 1.19 \\ \hline
        oBERTa\textsubscript{MEDIUM} Quantized & 103.47 & 6.20 & 9.65 & 9.60 & 0.57 \\ \hline
        oBERTa\textsubscript{MEDIUM} 90\%   & 106.01 & 6.35 & 9.42 & 9.34 & 0.28 \\ \hline
        oBERTa\textsubscript{MEDIUM} 90\% Quantized  & 149.25 & 8.94 & 6.69 & 6.65 & 0.42 \\ \hline
        oBERTa\textsubscript{SMALL} & 64.93 & 3.89 & 15.39 & 15.31 & 0.66 \\ \hline
        oBERTa\textsubscript{SMALL} Quantized & 208.09 & 12.47 & 4.80 & 4.78 & 0.28 \\ \hline
        oBERTa\textsubscript{SMALL} 90\% & 203.95 & 12.22 & 4.89 & 4.86 & 0.33 \\ \hline
        oBERTa\textsubscript{SMALL} 90\% Quantized & 270.63 & 16.21 & 3.69 & 3.68 & 0.25 \\ \hline
    \end{tabular}}
    \caption{Inference performance of the oBERTa model family using a batch size of 1, 24 cores, and a sequence length of 384}
    \label{tab:inference-bs-1-core24}
\end{table*}

\begin{table*}[!ht]
    \centering
    \scalebox{0.6}{
    \begin{tabular}{|l|l|l|l|l|l|}
    \hline
        model & Throughput (items/sec) & Speedup & Latency Mean (ms/batch) & Latency Median (ms/batch & Latency Std (ms/batch) \\ \hline
        oBERTa\textsubscript{base} & 19.55 & 1.00 & 818.23 & 811.93 & 15.52 \\ \hline
        oBERTa\textsubscript{base} Quantized & 83.92 & 4.29 & 190.65 & 189.55 & 4.21 \\ \hline
        oBERTa\textsubscript{base} 90\% & 74.29 & 3.80 & 215.35 & 214.31 & 2.47 \\ \hline
        oBERTa\textsubscript{base} 90\% Quantized & 137.83 & 7.05 & 116.07 & 115.43 & 2.56 \\ \hline
        oBERTa\textsubscript{base} 95\% & 89.07 & 4.56 & 179.62 & 178.92 & 3.19 \\ \hline
        oBERTa\textsubscript{base} 95\% Quantized & 160.68 & 8.22 & 99.56 & 98.91 & 2.63 \\ \hline
        oBERTa\textsubscript{MEDIUM} & 38.95 & 1.99 & 410.73 & 408.13 & 6.11 \\ \hline
        oBERTa\textsubscript{MEDIUM} Quantized & 157.12 & 8.04 & 101.82 & 101.27 & 2.21 \\ \hline
        oBERTa\textsubscript{MEDIUM} 90\%   & 144.95 & 7.41 & 110.37 & 109.62 & 1.56 \\ \hline
        oBERTa\textsubscript{MEDIUM} 90\% Quantized  & 251.32 & 12.86 & 63.65 & 63.40 & 1.76 \\ \hline
        oBERTa\textsubscript{SMALL} & 77.49 & 3.96 & 206.46 & 205.75 & 2.07 \\ \hline
        oBERTa\textsubscript{SMALL} Quantized & 276.10 & 14.12 & 57.94 & 57.43 & 1.63 \\ \hline
        oBERTa\textsubscript{SMALL} 90\% & 281.57 & 14.40 & 56.81 & 56.73 & 0.64 \\ \hline
        oBERTa\textsubscript{SMALL} 90\% Quantized & 417.35 & 21.35 & 38.32 & 38.01 & 1.55 \\ \hline
    \end{tabular}}
    \caption{Inference performance of the oBERTa model family using a batch size of 16, 24 cores, and a sequence length of 384}
    \label{tab:inference-bs-16-core24}
\end{table*}

\begin{table*}[!ht]
    \centering
    \scalebox{0.6}{
        \begin{tabular}{|l|l|l|l|l|l|}
    \hline
        model & Throughput (items/sec) & Speedup & Latency Mean (ms/batch) & Latency Median (ms/batch & Latency Std (ms/batch) \\ \hline
        oBERTa\textsubscript{base} & 19.02 & 1.00 & 3365.11 & 3352.63 & 29.49 \\ \hline
        oBERTa\textsubscript{base} Quantized & 84.80 & 4.46 & 754.73 & 749.38 & 18.69 \\ \hline
        oBERTa\textsubscript{base} 90\% & 72.22 & 3.80 & 886.13 & 881.75 & 10.65 \\ \hline
        oBERTa\textsubscript{base} 90\% Quantized & 140.14 & 7.37 & 456.67 & 453.59 & 11.03 \\ \hline
        oBERTa\textsubscript{base} 95\% & 88.35 & 4.64 & 724.41 & 720.43 & 10.85 \\ \hline
        oBERTa\textsubscript{base} 95\% Quantized & 162.76 & 8.56 & 393.21 & 390.45 & 12.15 \\ \hline
        oBERTa\textsubscript{MEDIUM} & 37.94 & 1.99 & 1686.85 & 1685.03 & 8.09 \\ \hline
        oBERTa\textsubscript{MEDIUM} Quantized & 160.48 & 8.44 & 398.80 & 396.47 & 9.27 \\ \hline
        oBERTa\textsubscript{MEDIUM} 90\%   & 130.02 & 6.84 & 492.22 & 486.90 & 9.64 \\ \hline
        oBERTa\textsubscript{MEDIUM} 90\% Quantized  & 259.51 & 13.64 & 246.61 & 244.54 & 7.13 \\ \hline
        oBERTa\textsubscript{SMALL} & 75.81 & 3.99 & 844.15 & 841.30 & 8.72 \\ \hline
        oBERTa\textsubscript{SMALL} Quantized & 267.70 & 14.07 & 239.06 & 237.86 & 7.02 \\ \hline
        oBERTa\textsubscript{SMALL} 90\% & 278.93 & 14.67 & 229.43 & 228.41 & 3.43 \\ \hline
        oBERTa\textsubscript{SMALL} 90\% Quantized & 455.71 & 23.96 & 140.43 & 139.81 & 5.40 \\ \hline
    \end{tabular}}
    \caption{Inference performance of the oBERTa model family using a batch size of 64, 24 cores, and a sequence length of 384}
    \label{tab:inference-bs-64-core24}
\end{table*}

\begin{table*}[!ht]
    \centering
    \scalebox{0.6}{
    \begin{tabular}{|l|l|l|l|l|l|}
    \hline
        model & Throughput (items/sec) & Speedup & Latency Mean (ms/batch) & Latency Median (ms/batch & Latency Std (ms/batch) \\ \hline
        oBERTa\textsubscript{base} & 4.89 & 1.00 & 204.65 & 204.93 & 1.82 \\ \hline
        oBERTa\textsubscript{base} Quantized & 20.01 & 4.09 & 49.95 & 49.88 & 0.66 \\ \hline
        oBERTa\textsubscript{base} 90\% & 17.60 & 3.60 & 56.82 & 56.70 & 0.72 \\ \hline
        oBERTa\textsubscript{base} 90\% Quantized & 37.50 & 7.67 & 26.66 & 26.61 & 0.38 \\ \hline
        oBERTa\textsubscript{base} 95\% & 20.15 & 4.12 & 49.62 & 49.60 & 0.54 \\ \hline
        oBERTa\textsubscript{base} 95\% Quantized & 46.02 & 9.41 & 21.72 & 21.70 & 0.31 \\ \hline
        oBERTa\textsubscript{MEDIUM} & 9.59 & 1.96 & 104.28 & 104.33 & 0.90 \\ \hline
        oBERTa\textsubscript{MEDIUM} Quantized & 41.23 & 8.43 & 24.25 & 24.18 & 0.33 \\ \hline
        oBERTa\textsubscript{MEDIUM} 90\%   & 38.30 & 7.83 & 26.10 & 26.05 & 0.41 \\ \hline
        oBERTa\textsubscript{MEDIUM} 90\% Quantized  & 73.28 & 14.99 & 13.64 & 13.60 & 0.19 \\ \hline
        oBERTa\textsubscript{SMALL} & 19.31 & 3.95 & 51.78 & 51.74 & 0.35 \\ \hline
        oBERTa\textsubscript{SMALL} Quantized & 75.81 & 15.50 & 13.18 & 13.18 & 0.19 \\ \hline
        oBERTa\textsubscript{SMALL} 90\% & 68.70 & 14.05 & 14.55 & 14.50 & 0.20 \\ \hline
        oBERTa\textsubscript{SMALL} 90\% Quantized & 145.57 & 29.77 & 6.86 & 6.86 & 0.11 \\ \hline
    \end{tabular}}
    \caption{Inference performance of the oBERTa model family using a batch size of 1, 4 cores, and a sequence length of 384}
    \label{tab:inference-bs-1-core4}
\end{table*}

\begin{table*}[!ht]
    \centering
    \scalebox{0.6}{
    \begin{tabular}{|l|l|l|l|l|l|}
    \hline
        model & Throughput (items/sec) & Speedup & Latency Mean (ms/batch) & Latency Median (ms/batch & Latency Std (ms/batch) \\ \hline
        oBERTa\textsubscript{base} & 5.14 & 1.00 & 3113.07 & 3113.92 & 19.89 \\ \hline
        oBERTa\textsubscript{base} Quantized & 22.14 & 4.31 & 722.72 & 719.24 & 11.40 \\ \hline
        oBERTa\textsubscript{base} 90\% & 17.15 & 3.34 & 932.97 & 931.21 & 5.76 \\ \hline
        oBERTa\textsubscript{base} 90\% Quantized & 39.03 & 7.59 & 409.90 & 408.71 & 4.64 \\ \hline
        oBERTa\textsubscript{base} 95\% & 19.80 & 3.85 & 808.16 & 806.80 & 4.15 \\ \hline
        oBERTa\textsubscript{base} 95\% Quantized & 46.54 & 9.06 & 343.75 & 342.75 & 4.12 \\ \hline
        oBERTa\textsubscript{MEDIUM} & 10.24 & 1.99 & 1563.00 & 1557.90 & 16.53 \\ \hline
        oBERTa\textsubscript{MEDIUM} Quantized & 42.82 & 8.33 & 373.61 & 372.88 & 4.05 \\ \hline
        oBERTa\textsubscript{MEDIUM} 90\%   & 33.69 & 6.56 & 474.88 & 474.25 & 3.64 \\ \hline
        oBERTa\textsubscript{MEDIUM} 90\% Quantized  & 76.10 & 14.81 & 210.24 & 209.41 & 2.45 \\ \hline
        oBERTa\textsubscript{SMALL} & 20.41 & 3.97 & 783.81 & 782.99 & 6.59 \\ \hline
        oBERTa\textsubscript{SMALL} Quantized & 79.57 & 15.48 & 201.07 & 200.60 & 2.12 \\ \hline
        oBERTa\textsubscript{SMALL} 90\% & 72.92 & 14.19 & 219.40 & 218.84 & 2.53 \\ \hline
        oBERTa\textsubscript{SMALL} 90\% Quantized & 139.50 & 27.14 & 114.68 & 114.45 & 1.53 \\ \hline
    \end{tabular}}
    \caption{Inference performance of the oBERTa model family using a batch size of 16, 4 cores, and a sequence length of 384}
    \label{tab:inference-bs-16-core4}
\end{table*}
\begin{table*}[!ht]
    \centering
    \scalebox{0.6}{
    \begin{tabular}{|l|l|l|l|l|l|}
    \hline
        model & Throughput (items/sec) & Speedup & Latency Mean (ms/batch) & Latency Median (ms/batch & Latency Std (ms/batch) \\ \hline
        oBERTa\textsubscript{base} & 5.06 & 1.00 & 12655.34 & 12680.81 & 57.78 \\ \hline
        oBERTa\textsubscript{base} Quantized & 21.88 & 4.32 & 2924.89 & 2921.95 & 31.78 \\ \hline
        oBERTa\textsubscript{base} 90\% & 17.18 & 3.40 & 3724.72 & 3724.23 & 15.27 \\ \hline
        oBERTa\textsubscript{base} 90\% Quantized & 37.44 & 7.40 & 1709.44 & 1699.64 & 26.97 \\ \hline
        oBERTa\textsubscript{base} 95\% & 22.13 & 4.37 & 2892.15 & 2893.08 & 22.94 \\ \hline
        oBERTa\textsubscript{base} 95\% Quantized & 43.94 & 8.68 & 1456.53 & 1451.76 & 20.45 \\ \hline
        oBERTa\textsubscript{MEDIUM} & 10.21 & 2.02 & 1567.70 & 1562.90 & 14.53 \\ \hline
        oBERTa\textsubscript{MEDIUM} Quantized & 42.74 & 8.45 & 374.35 & 373.15 & 4.00 \\ \hline
        oBERTa\textsubscript{MEDIUM} 90\%   & 33.99 & 6.72 & 470.67 & 469.99 & 3.58 \\ \hline
        oBERTa\textsubscript{MEDIUM} 90\% Quantized  & 75.64 & 14.95 & 211.53 & 210.80 & 2.61 \\ \hline
        oBERTa\textsubscript{SMALL} & 20.42 & 4.03 & 783.67 & 783.29 & 5.16 \\ \hline
        oBERTa\textsubscript{SMALL} Quantized & 79.44 & 15.70 & 201.40 & 201.43 & 2.90 \\ \hline
        oBERTa\textsubscript{SMALL} 90\% & 71.50 & 14.13 & 223.77 & 223.41 & 1.78 \\ \hline
        oBERTa\textsubscript{SMALL} 90\% Quantized & 139.55 & 27.58 & 114.65 & 114.48 & 1.53 \\ \hline
    \end{tabular}}
    \caption{Inference performance of the oBERTa model family using a batch size of 64, 4 cores, and a sequence length of 384}
    \label{tab:inference-bs-64-core4}
\end{table*}
\begin{table*}[!ht]
    \centering
    \scalebox{0.5}{
    \begin{tabular}{|l|l|l|l|l|l|l|}
    \hline
        Model & Throughput (items/sec) & Speedup vs BERT-Base & Speedup vs BERT-Large & Latency Mean (ms/batch) & Latency Median (ms/batch & Latency Std (ms/batch) \\ \hline
        bert\textsubscript{base} & 4.923 & 1.00 & 5.65 & 203.1165 & 202.7077 & 1.3646 \\ \hline
        bert-large & 0.8706 & 0.18 & 1.00 & 1148.6105 & 1145.145 & 9.5526 \\ \hline
        oBERTa\textsubscript{base} & 4.89 & 0.99 & 5.61 & 204.65 & 204.93 & 1.82 \\ \hline
        oBERTa\textsubscript{base} Quantized & 20.01 & 4.07 & 22.99 & 49.95 & 49.88 & 0.66 \\ \hline
        oBERTa\textsubscript{base} 90\% & 17.60 & 3.57 & 20.21 & 56.82 & 56.70 & 0.72 \\ \hline
        oBERTa\textsubscript{base} 90\% Quantized & 37.50 & 7.62 & 43.07 & 26.66 & 26.61 & 0.38 \\ \hline
        oBERTa\textsubscript{base} 95\% & 20.15 & 4.09 & 23.14 & 49.62 & 49.60 & 0.54 \\ \hline
        oBERTa\textsubscript{base} 95\% Quantized & 46.02 & 9.35 & 52.86 & 21.72 & 21.70 & 0.31 \\ \hline
        oBERTa\textsubscript{MEDIUM} & 9.59 & 1.95 & 11.01 & 104.28 & 104.33 & 0.90 \\ \hline
        oBERTa\textsubscript{MEDIUM} Quantized & 41.23 & 8.37 & 47.36 & 24.25 & 24.18 & 0.33 \\ \hline
        oBERTa\textsubscript{MEDIUM} 90\%   & 38.30 & 7.78 & 43.99 & 26.10 & 26.05 & 0.41 \\ \hline
        oBERTa\textsubscript{MEDIUM} 90\% Quantized  & 73.28 & 14.89 & 84.18 & 13.64 & 13.60 & 0.19 \\ \hline
        oBERTa\textsubscript{SMALL} & 19.31 & 3.92 & 22.18 & 51.78 & 51.74 & 0.35 \\ \hline
        oBERTa\textsubscript{SMALL} Quantized & 75.81 & 15.40 & 87.07 & 13.18 & 13.18 & 0.19 \\ \hline
        oBERTa\textsubscript{SMALL} 90\% & 68.70 & 13.95 & 78.91 & 14.55 & 14.50 & 0.20 \\ \hline
        oBERTa\textsubscript{SMALL} 90\% Quantized & 145.57 & 29.57 & 167.21 & 6.86 & 6.86 & 0.11 \\ \hline
        pruneOFA-large 80\% Quantized & 12.7315 & 2.59 & 14.62 & 78.5322 & 78.3961 & 0.4826 \\ \hline
        prunedOFA-large 90\% Quantized & 11.7265 & 2.38 & 13.47 & 85.2647 & 85.1616 & 0.4292 \\ \hline
        obert-large  & 0.876 & 0.18 & 1.01 & 1141.5707 & 1138.5756 & 9.0121 \\ \hline
        obert-large 95\% & 7.508 & 1.53 & 8.62 & 133.1785 & 132.9672 & 1.0091 \\ \hline
        obert-large 95\% Quantized & 16.8077 & 3.41 & 19.31 & 59.4828 & 59.322 & 0.6445 \\ \hline
        pruneBERT & 17.60 & 3.57 & 20.21 & 56.82 & 56.70 & 0.72 \\ \hline
        obert-large 97\% & 8.0414 & 1.63 & 9.24 & 124.3431 & 124.1421 & 1.0249 \\ \hline
        obert-large 97\% Quantized & 15.8631 & 3.22 & 18.22 & 63.0278 & 62.9979 & 0.6018 \\ \hline
        obert\textsubscript{base} 90\% & 18.2881 & 3.71 & 21.01 & 54.6688 & 54.5896 & 0.5476 \\ \hline
        obert\textsubscript{base} 90\% Quantized & 34.2797 & 6.96 & 39.37 & 29.1616 & 29.0977 & 0.3156 \\ \hline
        obert\textsubscript{base} 95\%  & 25.1818 & 5.12 & 28.92 & 39.6997 & 39.5986 & 0.5805 \\ \hline
        obert\textsubscript{base} 95\% Quantized & 40.6387 & 8.25 & 46.68 & 24.5986 & 24.5222 & 0.3231 \\ \hline
    \end{tabular}}
    \caption{Inference performance of the other sparse models using a batch size of 1, 4 cores, and a sequence length of 384 comparing the oBERTa models to previous sparse language models such as pruneOFA \cite{Zafrir2021PruneOF} PruneBERT \cite{Sanh2020MovementPA} and oBERT \cite{Kurti2022TheOB}}
    \label{tab:inference-bs-1-core4-competitive}
\end{table*}

\begin{table*}[!ht]
    \centering
    \scalebox{0.950}{
    \begin{tabular}{|l|l|*2l|*2l|}
    \toprule
        \multicolumn{2}{l}{} &  \multicolumn{2}{l}{Vs. BERT-Base} & \multicolumn{2}{l}{Vs. BERT-Large} \\ \hline
        Model & F1 & Recovery & Speed up & Recovery & Speed up \\ \hline
        BERT\textsubscript{base} & 88.55 & 100.00\% & 1.00 & 97.74\% & 5.65 \\ \hline
        BERT-large & 90.60 & 102.32\% & 0.18 & 100.00\% & 1.00 \\ \hline
        oBERTa\textsubscript{base} & 92.20 & 104.12\% & 0.99 & 101.77\% & 5.61 \\ \hline
        oBERTa\textsubscript{base} Quantized & 93.18 & 105.23\% & 4.07 & 102.85\% & 22.99 \\ \hline
        oBERTa\textsubscript{base} 90\% & 91.00 & 102.77\% & 3.57 & 100.44\% & 20.21 \\ \hline
        oBERTa\textsubscript{base} 90\% Quantized & 89.46 & 101.03\% & 7.62 & 98.74\% & 43.07 \\ \hline
        oBERTa\textsubscript{base} 95\% & 89.84 & 101.46\% & 4.09 & 99.16\% & 23.14 \\ \hline
        oBERTa\textsubscript{base} 95\% Quantized & 88.40 & 99.83\% & 9.35 & 97.57\% & 52.86 \\ \hline
        oBERTa\textsubscript{MEDIUM} & 90.36 & 102.04\% & 1.95 & 99.74\% & 11.01 \\ \hline
        oBERTa\textsubscript{MEDIUM} Quantized & 90.37 & 102.06\% & 8.37 & 99.75\% & 47.36 \\ \hline
        oBERTa\textsubscript{MEDIUM} 90\% & 89.26 & 100.80\% & 7.78 & 98.52\% & 43.99 \\ \hline
        oBERTa\textsubscript{MEDIUM} 90\% Quantized & 86.93 & 98.17\% & 14.89 & 95.95\% & 84.18 \\ \hline
        oBERTa\textsubscript{SMALL} & 84.87 & 95.84\% & 3.92 & 93.68\% & 22.18 \\ \hline
        oBERTa\textsubscript{SMALL} Quantized & 84.82 & 95.79\% & 15.40 & 93.62\% & 87.07 \\ \hline
        oBERTa\textsubscript{SMALL} 90\% & 84.66 & 95.61\% & 13.95 & 93.45\% & 78.91 \\ \hline
        oBERTa\textsubscript{SMALL} 90\% Quantized & 78.71 & 88.89\% & 29.57 & 86.88\% & 167.21 \\ \hline
        pruneOFA-large 80\% Quantized & 90.30 & 101.98\% & 2.59 & 99.67\% & 14.62 \\ \hline
        pruneOFA-large 90\% Quantized & 89.96 & 101.59\% & 2.38 & 99.29\% & 13.47 \\ \hline
        oBERT-large 95\% & 90.19 & 101.85\% & 1.53 & 99.55\% & 1.01 \\ \hline
        oBERT-large 95\% Quantized & 90.21 & 101.87\% & 3.41 & 99.57\% & 8.62 \\ \hline
        pruneBERT & 84.90 & 95.88\% & 3.41 & 93.71\% & 19.31 \\ \hline
        oBERT-large 97\% & 90.18 & 101.84\% & 13.05 & 99.54\% & 73.82 \\ \hline
        oBERT-large 97\% Quantized & 90.13 & 101.78\% & 1.63 & 99.48\% & 9.24 \\ \hline
        oBERT\textsubscript{base} 90\% & 88.47 & 99.91\% & 3.22 & 97.65\% & 18.22 \\ \hline
        oBERT\textsubscript{base} 90\% Quantized & 88.00 & 99.38\% & 3.71 & 97.13\% & 21.01 \\ \hline
        oBERT\textsubscript{base} 95\% & 88.19 & 99.59\% & 6.96 & 97.34\% & 39.37 \\ \hline
        oBERT\textsubscript{base} 95\% Quantized & 88.11 & 99.50\% & 5.12 & 97.25\% & 28.92 \\ \hline
        \bottomrule
    \end{tabular}}
    \caption{Speedups of the oBERTa-family as compared to existing published sparse models as compared to the performance of BERT\textsubscript{base} and BERT-large. Speedup measures the reduction in latency of MS/batch.}
    \label{tab:inference-competitive-full}
\end{table*}
\subsection{Limitations}
While much of our work has focused on showcasing the broad usability of compressed language models, they are not without fault. While our experiments focus on the compression of RoBERTa, the size of its training dataset makes complete exploration of the ability of pruning during pretraining somewhat limited. The work in the paper shows the ability to compress RoBERTa on a smaller pretraining dataset but does not contrast it with the impact of compression on the full dataset. \\
A second limitation of our work is the high computational demand required for creating public domain sparse language models. Despite amortizing the cost of compression to a few pretraining training regimes, the reduction of other language models like ALBERT \cite{Lan2019ALBERTAL} or XLM-R \cite{XLMR} require completely new training, pruning, and transfer experiments. 
\subsection{Responsible NLP Research - Reproducibility Checklist}
\subsubsection{Scientific Artifacts}
\noindent\textbf{Datasets.} We experiment with well-established benchmarks with usage in many broad domains. We do not perform any modification or augmentation in any dataset. Since datasets are not modified, we did not look for any personal or sensitive content. \\
In our pre-training experiments, we leverage the Toronto Book Corpus (TBC) \cite{Zhu2015AligningBA}\footnote{https://huggingface.co/datasets/bookcorpus} and the Wikipedia \cite{wikidump}\footnote{https://huggingface.co/datasets/wikipedia}. 
For fine-tuning we make use of SQuAD v1.1 \cite{Rajpurkar2016SQuAD1Q} \footnote{https://huggingface.co/datasets/squad}, SQuAD v2.0 \cite{Rajpurkar2018KnowWY} \footnote{https://huggingface.co/datasets/squadv2}, 
Quora Duplicate Question Dataset (QQP) \cite{Shankar2017IdentifyingQQ}\footnote{https://huggingface.co/datasets/glue}, and Multi-Genre Natural Language Inference (MNLI) \cite{N18-1101} \footnote{ https://huggingface.co/datasets/glue},  Large Movie Review Dataset (IMDB) \cite{maas-EtAl:2011:ACL-HLT2011}\footnote{ https://huggingface.co/datasets/imdb}, Stanford Sentiment Treebank (SST-2) \cite{socher-etal-2013-recursive}\footnote{ https://huggingface.co/datasets/glue}, and the shared task of CoNLL-2003 concerns language-independent named entity recognition (CONLL-2003) \cite{tjong-kim-sang-de-meulder-2003-introduction}\footnote{ https://huggingface.co/datasets/conll2003}datasets.

\noindent\textbf{Models.} The model used as a starting point for all of our experiments is RoBERta, publicly available via HuggingFace Hub~\footnote{https://huggingface.co/bert-base-uncased}. All other models presented in this paper will be released in openly-available repositories along with their compression recipes, training metrics, and hyper-parameters.

\subsubsection{Computational Experiments}
\textbf{Upstream.} During upstream pruning due to the large size of language models and their associated teachers we leverage 4x A100 40GB NVIDIA GPUs. We train for 5 epochs and an entire training and pruning run takes approximately 72 hours. Since the cost of such a large compute instance is high, these experiments were only run with a single seed and without major hyper-parameter exploration.

\textbf{Sparse-Transfer} Our experimentation on finetuning our compressed models uses the workhorse 16GB V100. Our sparse-transfer datasets vary greatly in size and as a result, so do experiments. Finetuning for CONL2003 takes less than 10 minutes while larger datasets like QQP take about 24 hours. Due to the number of datasets which we evaluate and the number of models in the oBERTa family, we only perform experimentation with a single fixed seed.

\noindent\textbf{DeepSparse inference.} We pair our compressed models with DeepSparse \cite{deepsparse} a publicly-available sparsity-aware CPU inference engine. All models are exported using the standard ONNX\footnote{https://onnx.ai/} format. For our competitive benchmarking against existing compressed language models, we leverage the model representations shared in the SparseZoo \footnote{https://sparsezoo.neuralmagic.com/}. This approach means that some older models such as oBERT may have had less optimized ONNA exports. We believe this difference in exportation causes the nearly 4x improvement in the performance of oBERTa base vs bert-base. 

\subsubsection{Computational Packages}
All of our experimentation is done using public libraries and datasets to ensure extensibility and reproducibility. Our experimentation is done using NeuralMagic's SparseML \footnote{https://github.com/neuralmagic/sparseml} which has specialized integration with HuggingFace's Transformers \footnote{https://github.com/huggingface/transformers} and Datasets \footnote{https://github.com/huggingface/datasets} libraries.